\pdfoutput=1
\documentclass{article}

\usepackage{PRIMEarxiv}

\usepackage[utf8]{inputenc} 
\usepackage[T1]{fontenc}    
\usepackage{hyperref}       
\usepackage{url}            
\usepackage{booktabs}       
\usepackage{amsfonts}       
\usepackage{nicefrac}       
\usepackage{microtype}      
\usepackage{latexsym}
\usepackage{graphicx}
\usepackage{multicol,multirow}
\usepackage{amsmath,amssymb,amsfonts}
\usepackage{mathrsfs}
\usepackage{amsthm}
\usepackage{rotating}
\usepackage{appendix}
\usepackage{ifpdf}
\usepackage[T1]{fontenc}
\usepackage{times}
\usepackage{newtxmath}
\usepackage{textcomp}%
\usepackage{xcolor}%
\usepackage{lipsum}
\usepackage{fancyhdr}       
\usepackage{graphicx}       
\graphicspath{{media/}}     
\newcommand{\ci}{\perp\!\!\!\perp}

\makeatletter
\newcommand{\settitle}{\@maketitle}
\makeatother

\hypersetup{
    colorlinks=true,
    linkcolor=black,
    citecolor=black,
    filecolor=black,
    urlcolor=black,
    hyperindex=true,
    bookmarks=true,
    bookmarksopen=true,
    bookmarksnumbered=true,
}

\title{Selecting Robust Features for Machine Learning Applications using Multidata Causal Discovery
}

\author{
  Saranya Ganesh S., Tom Beucler, Frederick Iat-Hin Tam, Milton S. Gomez \\
  Institute of Earth Surface Dynamics \\
  University of Lausanne \\
  Lausanne, Vaud, Switzerland\\
   \And
  Jakob Runge \\ 
  1. Institute of Data Science \\
  German Aerospace Center (DLR) \\
  Jena, Thuringia, Germany \\
  2. Faculty of Electrical Engineering and Computer Science\\
  Technische Universität Berlin \\
  Berlin, Germany \\
  \And
  Andreas Gerhardus \\
  Institute of Data Science \\
  German Aerospace Center (DLR) \\
  Jena, Thuringia, Germany\\
}

\begin{document}
\settitle

\begin{abstract}
Robust feature selection is vital for creating reliable and interpretable Machine Learning (ML) models. When designing statistical prediction models in cases where domain knowledge is limited and underlying interactions are unknown, choosing the optimal set of features is often difficult. To mitigate this issue, we introduce a Multidata (M) causal feature selection approach that simultaneously processes an ensemble of time series datasets and produces a single set of causal drivers. 
This approach uses the causal discovery algorithms PC$_1$ or PCMCI that are implemented in the \href{https://github.com/jakobrunge/tigramite}{Tigramite} Python package. These algorithms utilize conditional independence tests to infer parts of the causal graph. Our causal feature selection approach filters out causally-spurious links before passing the remaining causal features as inputs to ML models (Multiple linear regression, Random Forest) that predict the targets. We apply our framework to the statistical intensity prediction of Western Pacific Tropical Cyclones (TC), for which it is often difficult to accurately choose drivers and their dimensionality reduction (time lags, vertical levels, and area-averaging). Using more stringent significance thresholds in the conditional independence tests helps eliminate spurious causal relationships, thus helping the ML model generalize better to unseen TC cases. M-PC$_1$ with a reduced number of features outperforms M-PCMCI, non-causal ML, and other feature selection methods (lagged correlation, random), even slightly outperforming feature selection based on eXplainable Artificial Intelligence. The optimal causal drivers obtained from our causal feature selection help improve our understanding of underlying relationships and suggest new potential drivers of TC intensification.
\end{abstract}


\paragraph{Impact Statement}: {\normalfont While causal feature selection helps design more robust ML models, its joint application to multiple dataset remains limited because standard causal discovery algorithms output a different set of drivers for each dataset, which is impractical. To mitigate this issue, we apply a newly-developed ``multidata'' causal feature selection approach, which identifies a single set of optimal causal drivers from an ensemble of multivariate time series. Applied to the statistical prediction of TC intensity, our approach outperforms standard feature selection methods by helping simple regression algorithms better generalize to unseen cases. In addition to making our models robust, causal feature selection also eliminates redundant predictors while identifying new ones, leading to lighter models and aiding scientific discovery.}

\section{Introduction}
Machine Learning (ML) combines statistical methods and numerical optimization to learn a group of tasks from data. Progress in computational capabilities, combined with the availability of large amounts of data, allows the development of ML models to predict and understand nonlinear systems such as climate processes and extreme weather events. For environmental applications, processing big data that are nonlinearly related often requires (i) dimensionality reduction; and (ii) strategically selecting the model's features to make ML models cheaper to run, generalize better, and easier to explain \cite{guyon2003introduction,yu2020causality,yu2022causal}. This Article compares different methods to discover a subset of the most relevant features in environmental datasets \cite{guyon2003introduction,li2017feature,post2016does} and explores the effect of causal feature selection on statistical prediction skill. For this, we work at the intersection of causal inference and ML, an active area of research \cite{chen2020causalml} because causal relations help acquire robust knowledge beyond the support of observed data distributions \cite{scholkopf2021toward}. Causal inference can broadly be categorized into three research directions: (i) causal representation learning; (ii) causal discovery; and (iii) causal reasoning \cite{scholkopf2021toward,kaddour2022causal}. To select features, we here explore the use of \textit{causal discovery}, a methodology for learning qualitative cause-and-effect relationships between a collection of variables from data that have not been obtained under controlled experimental conditions \cite{spirtes2000causation, peters2017elements}. Incorporating causal relationships in ML models via feature selection can make ML models more interpretable \cite{guyon2003introduction,runge2015identifying,yu2022causal,iglesias2023causally} and less susceptible to overfitting \cite{aliferis2010local1, aliferis2010local2,runge2015optimal}. 

There are two main challenges when applying causal discovery in environmental sciences. The first challenge is algorithmic: Often environmental data consists of multiple realizations of the same process with slight differences, and causal discovery algorithms that apply to such multiple realization problems remain underexploited \cite{yu2020causality}. The second challenge is the lack of benchmarking: Causal feature selection is rarely compared against other feature selection methods for ML-based predictions. Here, we address these two gaps by introducing a causal feature selection framework to estimate causal relationships from multiple time series datasets \cite{runge2015optimal,runge2015identifying,runge2019inferring,runge2019detecting,runge2020discovering, yu2019multi}. We compare feature selection algorithms by training simple ML prediction models for each of the selected sets of features and evaluating their predictive performances. Our framework is applied to the prediction of Tropical Cyclone (TC) intensity to demonstrate that causal feature selection (i) improves the out-of-sample skill, and (ii) uncovers the best predictors in real-world situations. 

\section{Methodology: Causal Feature Selection for Multiple Realizations\label{sub:Causal_Feature_Selection}}

Our implementation of causal feature selection \cite{gao2016efficient,geiger1990identifying,pena2007towards} uses the recently-developed Multidata (M) functionality for two causal discovery algorithms based on time-series, explained below. 
Our \emph{multidata} causal discovery approach used to pre-select causally relevant predictors has two steps: (i) the causal discovery algorithms; and (ii) applying these algorithms to a dataset comprising data from multiple sources. From a causal perspective, the setup used in this study is simplified because only the variables that are time-lagged with respect to the target variables are considered potential predictors. As a result, causal discovery effectively reduces to a feature selection algorithm that removes all those predictors which are (conditionally) independent of the target (given the other predictors) and which hence do not provide any additional information for predicting the target. The multidata functionality itself, however, is more general and also applies to the time series causal discovery tasks that also consider contemporaneous causal relationships. 

Here, we explore the use of the causal discovery algorithms PC$_1$ and PCMCI \cite{runge2019detecting} for causal feature selection. The \textbf{PC$_1$} algorithm is a variant of the PC algorithm \cite{spirtes2000causation}. First,  PC$_1$ initializes the potential causal drivers $pa(Y_t)$ of each target variable $Y_t$ as the set of all variables $pa(Y_t)=\{X^i_{t-\tau} ~\vert~ i = 1, \dots, N_X, \, \tau = \tau_{min}, \ldots, \tau_{\text{max}}\}$ within the considered range $[\tau_{min}, \tau_{max}]$ of time lags, where the $X^i$ with $i =1, \ldots, N_X$ are the potential predictors and where $\tau_{min}$ and $\tau_{max}$ respectively are the minimal and maximal time lags at which direct causal influences can occur.. Then, PC$_1$ iteratively removes variables from $pa(Y_t)$ that are irrelevant or redundant for the prediction of $Y_t$. Specifically, PC$_1$ removes elements $X^i_{t-\tau}$ from $pa(Y_t)$ that are conditionally independent of  $Y_t$ given subsets $S_k\subseteq pa(Y_t)$ whose cardinality $k$ increases iteratively. For $k=0$ all $X^i_{t-\tau}$ with $X^i_{t-\tau} \ci Y_t $ are removed, for $k=1$ those with $X^i_{t-\tau} \ci Y_t | S_1$ where $S_1$ is the strongest driver from the previous step, for $k=2$ those with $X^i_{t-\tau} \ci Y_t | S_2$ where $S_2$ are the two strongest drivers from the previous step, and so on. In this work, conditional independence is tested using partial correlation (in general, though, the algorithm can be combined with any conditional independence test). The corresponding independence test is based on a standard significance level $pc_{\alpha} = 0.02$ and uses a strength of association that is based on the absolute partial correlation value. This iteration is continued until $k$ is greater than the cardinality of $pa(Y_t)$. The PC algorithm is different from PC$_1$ in so far as that PC, for every $k$, does not only test for conditional independence given exactly one cardinality $k$ subset of $pa(Y_t)$ but tests for conditional independence given all cardinality $k$ subsets of $pa(Y_t)$.

The \textbf{PCMCI} algorithm \cite{runge2019detecting}, after first running PC$_1$, reinitializes all links and then subjects all links to the momentary conditional independence (MCI) tests $X^i_{t-\tau} \ci Y_t | pa(Y_t)\setminus \{X^i_{t-\tau}\}, pa(X^i_{t-\tau})$, removing the link if independence is not rejected. Here, the condition on $pa(X^i_{t-\tau})$ helps to remove false positives that tend to be inflated due to autocorrelation. Controlling false positives is important for a causal discovery setting but is not necessarily important for a causal feature selection/prediction setting as considered in this paper. Within the study presented here, we employ both the PC$_1$ and the PCMCI algorithm to empirically analyze which of the two methods is preferable for causal feature selection. As with all causal discovery methods, PC$_1$ and PCMCI rely on certain assumptions. The essential assumption is that conditional independencies in the data are in one-to-one correspondence with $d$-separations \cite{pearl1988probabilistic} in the causal graph \cite{verma1990causal, geiger1990identifying, spirtes2000causation, pearl2009causality}. Moreover, both methods assume \emph{causal sufficiency} \cite{spirtes2000causation}, i.e., the absence of unobserved variables that causally influence two observed variables. The latter assumption is not necessarily fulfilled in our context, even though we included a range of potential predictors. This means that some of the obtained causal features might still be spurious and may not work if the target distribution differs from the training distribution (out-of-distribution prediction).

When PC$_1$ and PCMCI are applied to a \textit{single} multivariate time series, 
samples are drawn from this time series in a sliding-window fashion. The drawn set is internally passed to the statistical hypothesis tests of conditional independencies. For this sliding-window approach to be valid, the causal relationships need to remain unchanged throughout the time series (\textit{causal stationarity} assumption). When PC$_1$ and PCMCI are applied to \textit{multiple} multivariate time series, if all time series of this ensemble can be assumed to share the same causal relationships within specific time ranges, then we can combine the sample sets from all ensemble members\footnote{each of the sample sets is obtained in a sliding window fashion from one of the member time series} into a single, larger dataset. This larger dataset, which includes data from multiple sources (e.g., from multiple storms), is then internally passed to the conditional independence tests. The PC$_1$ and PCMCI algorithms can then proceed as usual. Consequently, although the input is an ensemble of multivariate time series, the output is still a single set of predictors. In addition to its practicality, our \emph{multidata} approach benefits from an enlarged set of samples, increasing the power of the conditional independence tests. Hence, we expect the sets of predictors obtained by running multidata causal discovery on an ensemble of time series to be more reliable than the sets of predictors obtained by running causal discovery on any single member of this ensemble---if the assumption of a common causal structure holds. An alternative approach would be to run PC$_1$ and PCMCI on any single member time series and then appropriately aggregate the resulting sets of predictors (across the members). 
 
\section{Application: Statistical Prediction of Tropical Cyclone Intensity}

\subsection{Motivation}

The increasing frequency of intense TCs \cite{emanuel2005increasing,knutson2020tropical} combined with growing coastal populations have escalated the vulnerability of the tropical urban coasts. For context, more than half of Earth's population is projected to live in the Tropics by 2050 \cite{edelman2014state} and more than a billion people worldwide could be living in low-elevation coastal zones by 2060, particularly in South and East Asia \cite{neumann2015future}. Predicting storm intensity changes, including rapid intensification in TCs, remains a major challenge \cite{demaria2014tropical}, because of unresolved complexities of storm dynamics in numerical models. Furthermore, numerical models suffer from a reduction in forecast skills with an increase in lead time \cite{saranya2018new}.An alternative to numerical forecasting is statistical forecasting, as statistical models can improve cyclogenesis and intensity forecasts \cite{chen2020machine,kim2019machine}. For instance, statistical models based on logistic regression, random forest, decision tree, and randomized decision trees \cite{su2020applying} outperformed the National Hurricane Center in predicting TC rapid intensifications over the Atlantic and Eastern Pacific basins \cite{rozoff2011new,kaplan2010revised}. A potential drawback of statistical models is that it is often difficult to choose appropriate predictors for reliable forecasts. To better predict TC intensity, the models need to represent the physical mechanisms behind TC intensification more accurately; these include large-scale circulations, local conditions, and internal processes \cite{emanuel2016predictability,kaplan2010revised}. We argue that one way to make statistical models more robust is to apply causal discovery algorithms and eliminate causally-spurious predictors. In this study, we apply the PC$_1$ and PCMCI methods to generate a single set of causally relevant predictors from multiple TC time series. 

\subsection{Data}

The TC dataset is created using multiple environmental variables at different pressure levels known to be favorable for TC intensification \cite{sikora1976investigation, petty2000improving, li2011revisiting} from the global high-resolution ECMWF ReAnalysis-5 \cite{hersbach2020era5}(ERA5) with 25 km horizontal resolution, and 3-hourly temporal resolution (see SI - section A for the full list). Here, we selected a total of 260 TC cases spanning from 2001 to 2020 in the Northwest Pacific basin (WPAC). The TCs with a lifetime of more than 6 days up to landfall are selected for the study to understand the effect of environmental parameters on TC intensity up to 3 days time lag, so each case has a time span from genesis up to landfall based on each TC best track. TC tracks are obtained from the International Best Track Archive for Climate Stewardship \cite{knapp2010international}. Rather than directly feeding the time series of predictor variables for the cases at each gridpoint around the storm, 
the values are averaged in horizontal areas defined with respect to the TC Center. Each atmospheric predictor is post-processed into two sets of time series representing inner-core (TC center to a radius of 200 km) and outer-core characteristics (annulus from a radius of 200-800 km). The choice of averaged areas follows the current practice in operational statistical intensity prediction schemes \cite{demaria2005further}. The distinction between outer-core and inner-core processes is justified because TC intensity is affected by environmental conditions in the storm's neighborhood and internal processes within the storm \cite{hendricks2019summary, sitkowski2009low}. From an ML perspective, this preliminary dimensionality reduction removes features with high spatial correlations, reduces the complexity of the statistical models, and possibly improves model generalizability by removing some of the predictors' spatial heterogeneity in different storms.
\\
Once this preliminary dimensionality reduction is done, our goal is to eliminate spurious \textit{features}, here defined based on meteorological variable, vertical (pressure) levels, time lag, and horizontal averaging sector (inner or outer core). We describe each TC using a total of $N_X = 234$ time series of horizontally-averaged 3D variables at given vertical levels and 2D variables (see SI Table 1 for details). With regards to the time lags, we explore the time steps between 24h before the target (corresponding to $\tau_{min} = 8$) and 72h before the target (corresponding to $\tau_{max} = 24$). This results in a total number of 3978 (234 potential predictors $\times $ 17 time steps) for the causal algorithms, which eliminate the spurious links between the features and the targets. 
We randomly split the data \textit{by TC} to avoid spatiotemporal correlation: Out of the selected 260 TCs, we randomly split 205 cases from 2001-2020 into a training set (150 cases) and validation set (55 cases) while keeping 55 cases from recent years (2017-2020) in the test set, without any overlaps. The regression task is to forecast three variables with a 1-day lead time, including (1) Maximum wind speed at 10m (Max. 10m Wind, in m/s), (2) Minimum Sea-Level Pressure (MSLP, in hPa), and (3) horizontally-integrated total precipitation (Tot. Intg. Precip. in $km^{2} $) accumulated over 3-hrly intervals. Maximum sustained wind speed at 10m (averaged over 1min, 3min, or 10min depending on the Regional Specialized Meteorological Centre) is the standard measurement for the intensity currently used operationally. We include MSLP as it correlates better with TC damage \cite{atkinson1977tropical,kaplan2010revised}. Additionally, MSLP is easier to estimate as it is an integrated quantity and only requires a couple of measurements near the storm center. Finally, we included total integrated precipitation as a potential target because most fatalities and damage from TCs are caused by heavy precipitation and storm surges \cite{rappaport2014fatalities}.

\begin{figure}[htb]%
    \centering
    \includegraphics[width=1.0\textwidth]{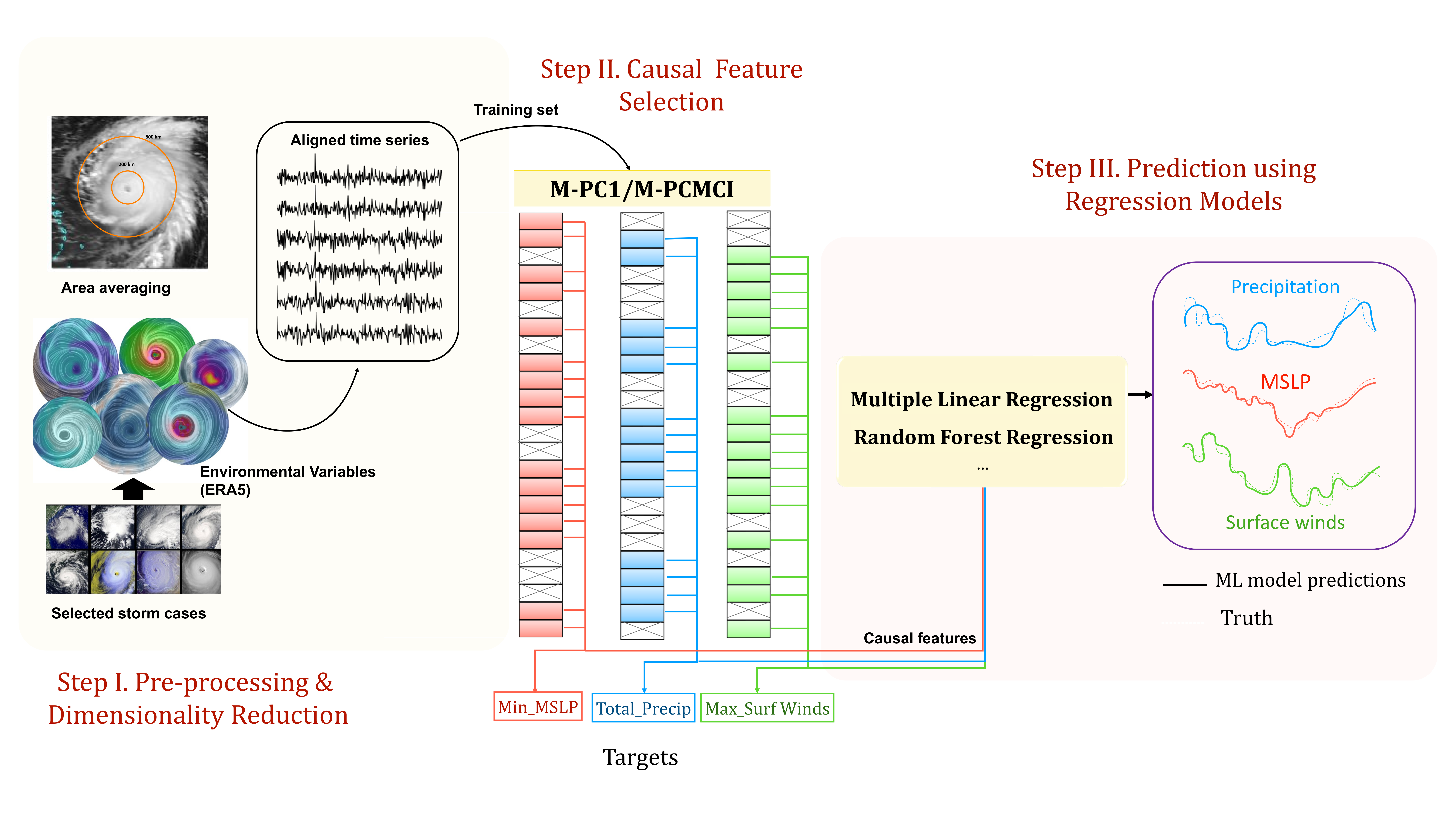}
    \caption{Multidata Causal Feature Selection applied to TC prediction: After reducing the dimensionality of spatiotemporal fields to yield time series for several TC cases (Step I), the ensemble of \textit{aligned} time series is fed to the multidata causal discovery algorithm to calculate the optimal set of causal drivers (Step II), which can be fed to a regression algorithm to make robust predictions (Step III)}
    \label{fig1}
\end{figure}
 
\subsection{Causal Machine Learning\label{sub:causal_ML}}
In this section, we describe the feature selection methodology in the context of TC intensity prediction. Our causal feature selection framework is shown in Fig~\ref{fig1}. Once the four-dimensional fields have been reduced to time series, we align the time series in the training set based on the minimum pressure value recorded during each TC's lifetime, which is a smoother measure of TC intensity than maximum surface wind speeds \cite{chavas2017physical}. Temporally aligning the multivariate time series of different ensemble members is key, as the resulting ensemble is more likely to satisfy the common causal structure assumption, improving prediction skills using causal feature selection. \\
After aligning the time series, we feed the training set as inputs to the PC$_1$ and PCMCI algorithms (both implemented in \href{https://github.com/jakobrunge/tigramite}{tigramite}) to extract the most significant causal features. Here, an input feature may be defined as an environmental or derived variable (see SI section A, Table S1) at any given pressure level which is spatially averaged either by the inner or outer core radii at a given 3 hourly time-step. We stress that PC$_1$ and PCMCI are only applied to the training data.
The implementation of both PC$_1$ and PCMCI contains several hyperparameters, including minimum and maximum time lags for the analysis, fixed to 1 day (8 timesteps) and 3 days (24 timesteps), respectively, for our prediction task. Further tunable hyperparameters are the significance level for conditional independence testing ($pc_{\alpha}$) and a significance threshold for the p-value matrix (alpha-level, only used for PCMCI), which control the selected number of features.
\\
Once PC$_1$ and PCMCI have selected the most significant causal drivers of the targets from the ensemble of time series, these drivers are used as inputs to the prediction model. We logarithmically vary the values of $pc_{\alpha}$ and alpha-level, which in effect controls the number of selected features, as more stringent ($pc_{\alpha}$) values will result in the selection of fewer and more significant features. From this set of experiments, we determine the best hyperparameters suitable for each target of interest by maximizing the validation performance of the trained regression models. We use Multiple linear regression (MLR) and Random Forest (RF) Regression models to predict the targets from the causally selected features.
 The MLR algorithms for Causal-MLR experiments were prepared using the Scikit Learn \cite{pedregosa2011scikit} implementation of the Linear Regression algorithm and its corresponding default parameters. Each MLR algorithm was trained to predict one of three unscaled target variables using the selected, standard-scaled features. We also included RF regression models using the same causal feature selection set-up (Fig~\ref{fig1}) to explore the impact of causal feature selection for more complex nonlinear regression methods. We used the RF regressor from the scikit learn library \cite{pedregosa2011scikit} to prepare the causal-RF and optimized its hyperparameters with a randomized search.

\subsection{Non-Causal Machine Learning Baselines\label{sub:noncausal_ML}}
This study is motivated by the working hypothesis that regression models using causally-selected features outperform non-causal baselines in terms of generalization. Here, \textit{non-causal baselines} subsume both the case of no feature selection and the case of feature selection based on non-causal criteria. 
We compare our causal feature selection to non-causal feature selection methods such as lagged correlation, random selection as well as an eXplainable Artificial Intelligence (XAI) method of feature selection using RF regression (More details are provided in SI section C.). To test our causal approach's ability to effectively use time lags, we also train a Long Short-Term Memory (LSTM) network using all time lags between $\tau_{min} $ and $\tau_{max} $ and without feature selection. We implement the LSTM using the PyTorch \cite{pytorch} library and conducted a hyperparameter search with the Optuna \cite{optuna_2019} framework. A more detailed description of the architecture is provided in SI Section C. 
\\
 \begin{figure}[htb]%
 \centering
 \includegraphics[width=1.0\textwidth]{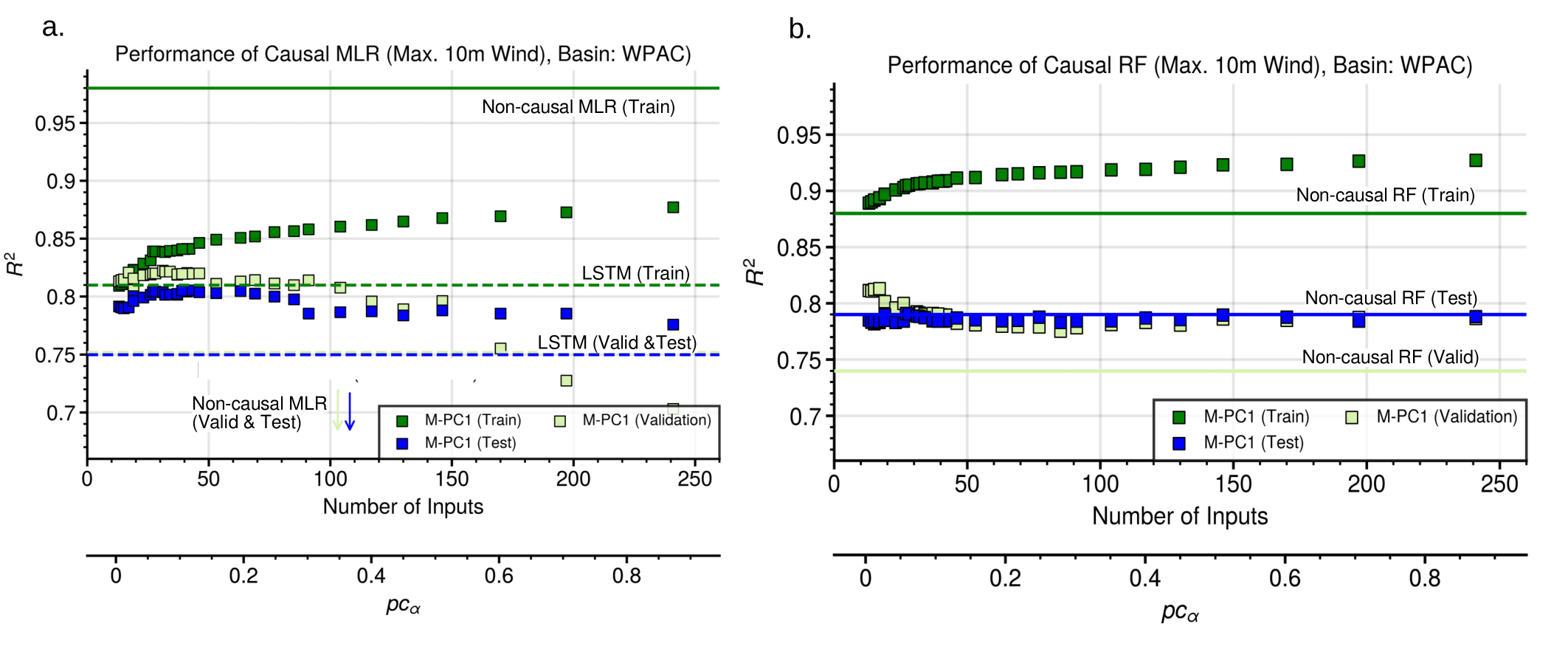}
 {\caption{(a.) Causal MLR models using M-PC$_1$ systematically outperform LSTMs (dashed line) on all sets and their non-causal counterparts (solid lines) on the validation and test sets; (b.) Causal RF models outperform their non-causal counterparts (solid lines) on the training and validation sets}
 \label{fig2}}
 \end{figure}
 
\section{Results}

\subsection{Performance of Causal Machine Learning}
Our first objective is to find the set of causal features that are best linked to the intensification in TCs at a lead time of 1-day. To measure the suitability of a set of causal features, we evaluate how MLR as well as RF models trained with the causally-selected features perform when predicting TCs that are unseen during training. We evaluate prediction skill holistically (see SI section D), but focus on the coefficient of determination ($R^2$) in the main text, with $R^2=1$ corresponding to a perfect prediction and $R^2=0$ to an error of one standard deviation. In Fig~\ref{fig2}, we show the performance of Causal-ML models to predict the maximum winds, 24 hours in advance using M-PC$_1$ method. A less stringent significance threshold results in a larger set of features that are retained during training, which has a clear negative effect on the model validation performance. We found that feature selection using M-PCMCI is comparable to M-PC$_1$ when the minimum time lag is 6 hours (shown in SI section B), but the performance of M-PCMCI drastically deteriorates when we increase the minimum time lag to 1 day.  Here, we only show the causal ML results based on M-PC$_1$ here. For comparison, similar experiments with reduced lead times where minimum and maximum time lags are set to 6 hours and 2 days, respectively, are shown in SI Section B. PCMCI's main advantage is to better control false positives in the presence of strong autocorrelation~\cite{runge2019detecting}, which is more important for an actual causal discovery setting than for the causally-informed prediction setting considered in this paper.

Causal MLR scores are better than those of non-causal MLRs (Fig~\ref{fig2}a), which use all inputs without feature selection. 
The non-causal baselines clearly overfit the training set. The Causal MLR is also compared with a recurrent neural network using an LSTM layer, and Causal MLRs outperform the best LSTM model for all targets, which is remarkable given their simplicity. When comparing the RF models, we find that Causal-RF scores are better than non-Causal-RF for the validation set, while test set scores are comparable for wind speed predictions. In general, the ($R^2$) values are highest for the predictions of MSLP, followed by wind and total integrated precipitation (Fig \ref{fig2}, S1, S2). The optimal set of causal drivers that performs well on the training and validation sets (without leading to overfitting) is sparse, containing only 31 features in the causal MLR case for predicting maximum wind (Table \ref{tab:R21}). This result suggests that many of the features are spuriously linked to TC intensity and can be removed without sacrificing the predictive skill of simple MLR models compared to a non-causal RF baseline. Similar results are obtained for the prediction of other targets, as shown in Fig S1, S2 in SI section B.

\subsection{Comparison with Feature Selection Baselines}
\begin{figure}[htb]%
    \centering
    \includegraphics[width=\textwidth]{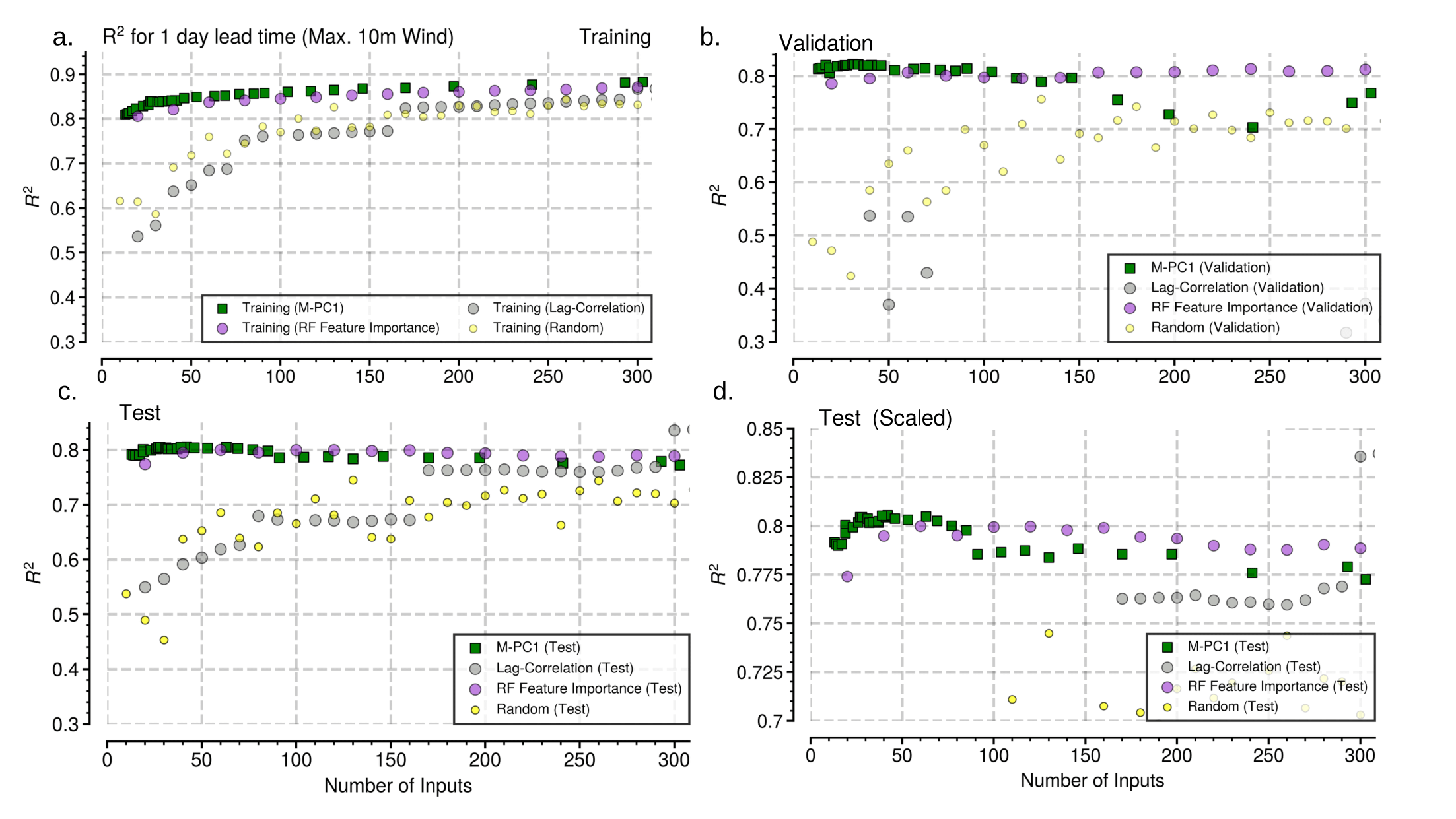}
    \caption{While (a) both causal and non-causal models fit the training set better when their number of features is increased, M-PC$_1$ causal feature selection provides the best generalization to unseen cases in the validation (b) and test sets (c and d zoomed-in version), especially when the number of input features is below 100 (d). For all methods, selected features are fed to MLR for predicting maximum winds for WPAC TCs at a lead time of 1 day}
    \label{fig3}
 \end{figure}

Our second objective is to compare the performance of Causal MLRs to MLRs with non-causal feature selection baselines (described in SI section C). For the maximum wind predictions (Fig~\ref{fig3}), PC$_1$ consistently outperforms the two simpler feature selection baselines (random and lagged correlation), especially on the validation and test sets (Fig~\ref{fig3}b,c). Lagged correlation, in particular, selects sets of predictors that perform very poorly in comparison, especially during validation. The ability of an XAI-based feature selection to capture nonlinearities seems to improve predictor selection, resulting in $R^2$ values that are almost comparable to the PC$_1$ causal feature selection method. Nevertheless, the causal PC$_1$ method retains an advantage for very sparse models (less than 50 input features), suggesting that the initial selection of causally-relevant predictors allows these sparse models to beat the corresponding noncausal sparse models. PC$_1$ performs better than the other methods for the two other targets, which is shown in SI section C (Fig~S3-S6). We note that lagged correlation performance is comparable to Causal-MLR and XAI-based feature selection in predicting Total integrated precipitation. This motivates adapting our conditional independence tests for non-normal distributions, which we leave for future work. The performance comparison based on ($R^2$) of all best ML models used in the study, along with their number of input features, is listed in Table \ref{tab:R21}.

\begin{table}
\begin{centering}
\resizebox{1.0\textwidth}{!}{%
\begin{tabular}
{c|c|c|c|c|c|c|c|c|c|c}
\multicolumn{2}{c|}{  } & \multicolumn{3}{c|}{Training {{} }\textit{{(No. of features)}}} & \multicolumn{3}{c|}{Validation} & \multicolumn{3}{c}{Test }\tabularnewline
\multicolumn{2}{c|}{ML Models / Target} & Pmin ($hPa$){{} } & V10 ($ms^{-1}$) & Precip $\times 10^{-3}$ ($km^{2}$) & Pmin{{} } & V10 & Precip & Pmin{{} } & V10 & Precip \tabularnewline
\hline 
\hline
\multicolumn{2}{c|}{\textbf{Causal-RF}} & 0.93 \textit{{(26)}} & 0.89 \textit{{(17)}} & 0.83 \textit{{(123)}} & 0.87 & 0.81 & 0.65 & 0.88 & 0.78 & 0.62\tabularnewline
\multicolumn{2}{c|}{\textbf{Causal-MLR}} & \textbf{0.87}\textit{{(17)}} & \textbf{0.84} \textit{{(31)}} & \textbf{0.71} \textit{{(90)}} & \textbf{0.88} & \textbf{0.82} & \textbf{0.68} & \textbf{0.89} & \textbf{0.80} & \textbf{0.62}\tabularnewline
\hline 
\multirow{3}{*}{\textbf{Non-Causal-RF}} & All & 0.93 \textit{{(3978)}} & 0.88 \textit{{(3978)}} & 0.75 \textit{{(3978)}} & 0.77 & 0.74 & 0.65 & 0.89 & 0.79 & 0.58\tabularnewline
 & Lagged & 0.96 \textit{{(480)}} & 0.93 \textit{{(560)}} & 0.79 \textit{{(80)}} & 0.85 & 0.81 & 0.69 & 0.89 & 0.81 & 0.61\tabularnewline
 & Random & 0.96 \textit{{(870)}} & 0.93 \textit{{(770)}} & 0.85 \textit{{(970)}} & 0.79 & 0.77 & 0.59 & 0.87 & 0.77 & 0.56\tabularnewline
\hline 
\multirow{4}{*}{\textbf{Non-Causal-MLR}} & All & 0.99 \textit{{(3978)}} & 0.98 \textit{{(3978)}} & 0.96 \textit{{(3978)}} & -0.94 & -10.85 & -127.98 & 0.51 & -0.01 & -0.39\tabularnewline
 & Lagged & 0.92 \textit{{(440)}} & 0.64 \textit{{(40)}} & 0.68 \textit{{(120)}} & 0.84 & 0.54 & 0.65 & 0.92 & 0.59 & 0.64\tabularnewline
 & Random & 0.91\textit{{{} (420)}} & 0.83 \textit{{(130)}} & 0.69 \textit{{(290)}} & 0.78 & 0.76 & 0.62 & 0.86 & 0.75 & 0.54\tabularnewline
 & XAI & 0.92 \textit{{(240)}} & 0.88 \textit{{(420)}} & 0.70 \textit{{(140)}} & 0.84 & 0.82 & 0.69 & 0.91 & 0.79 & 0.63\tabularnewline
\hline
\multicolumn{2}{c|}{\textbf{LSTM}} & 0.87 \textit{{(3978)}} & 0.81 \textit{{(3978)}} & 0.71 \textit{{(3978)}} & 0.77 & 0.75 & 0.65 & 0.81 & 0.75 & 0.61\tabularnewline
\hline
\end{tabular}}
\caption{$R^2$ score for each experiment's best model on the validation set, along with the number of selected features (in parentheses). \textbf{Causal-MLR} gives the best performance with the least features.\label{tab:R21}}
\par\end{centering}
\end{table}

\subsection{Optimal Causal Features}
\begin{figure}[htb]%
    \centering
    \includegraphics[width=\textwidth]{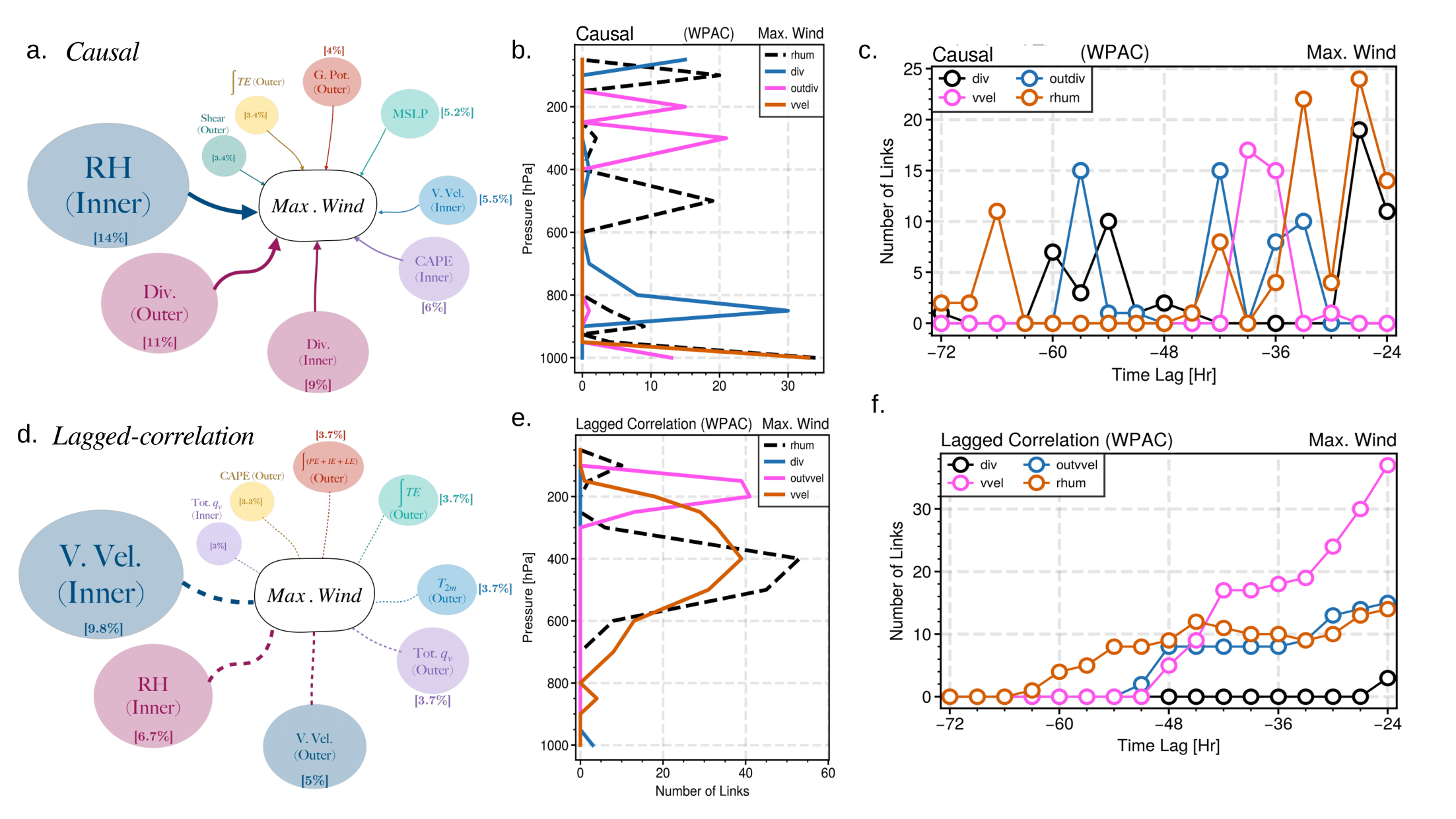}
    \caption{Most frequently and significant predictors used by the best Causal-MLR model organized by (a) top nine meteorological variables; (b) pressure level; and (c) time lag. (d-f) Most frequently selected features for the lag correlation method. For the two rightmost columns, we retained the four most frequent features (Relative humidity (inner), Vertical velocity (inner) and horizontal divergence (inner and outer).}
    \label{fig4}
 \end{figure}

Here, we expand upon our results from the previous section to understand \textit{why} causal-MLR models outperform the models that use other feature selection methods. For this purpose, we rely on the frequency of predictor selection (across the models)by the best-performing causal-MLR and lagged correlation MLR models\footnote{Best causal-MLR models are defined as model with $R^2$ within 1\% of the best validation $R^2$. This threshold is relaxed to 10\% for lagged correlation models to sample a comparable number of features.}. We find that both methods choose different predictors for the maximum wind predictions while identifying inner core relative humidity as critical for wind prediction. However, divergence is a major predictor in the causal-MLR (Fig~\ref{fig4}a.), despite not being in the 10 most frequently selected predictors for the lagged-correlation models (Fig \ref{fig4}d.). 
The vertical distribution (Fig \ref{fig4}b,e) and the time lag information (Fig~\ref{fig4}c,f) of the most frequently chosen features reveal several differences in the causal models as compared to the lagged correlation models. Unlike the lag correlation method, the causal method selects features at specific vertical levels and time lags that are most informative to the prediction (Fig~\ref{fig4}b) rather than placing importance over a wide range of vertical levels(Fig \ref{fig4}e) and lags. The PC$_1$ algorithm iteratively removes variables from the parent set $pa(Y_t)$ that are irrelevant or redundant for the prediction of the target, $Y_t$ via conditional independence tests (here, based on partial correlation). PC$_1$ then ranks features based on significance test statistics, which gives a good measure of predictor relevance. Once $X_{t}$ is in the parent set $pa(Y_t)$, its neighbors will be iteratively removed because they are not conditionally independent of $X_{t}$. This confirms the interpretation that the selected time lags and vertical levels are ``most predictive'' of $Y_t $, and that the spatiotemporal neighbors of $X_{t}$ are eliminated because of redundancy, which is due to the high spatiotemporal correlations in our dataset.  Next, from a scientific viewpoint, the causal models clearly emphasize the low-level inner-core convergence (div), middle and upper tropospheric relative humidity (RH, rhum)in the inner core and the upper-level divergence (outdiv) in the outer core as most important predictors whereas the lagged correlation models rely on middle- and upper-tropospheric vertical motions (vvel) for the prediction task. Finally, in the time-lag plots (Fig \ref{fig4}c,f), the divergence links in the causal models are chosen at time lags of more than 2 days (-60hr$<$t$<$-50hr), while lagged correlation models focus on features at the shortest lead times (t$>$-48hr) as they are more correlated with decreasing time lags.

Causal-MLR models rely on low-level convergence and upper-level divergence at longer time lags. In contrast, the lagged correlation MLR models mostly rely on mid- to upper-tropospheric vertical motion at shorter lead times. One way to interpret this is that the mid-tropospheric vertical motion could be a confounder, which is removed by the PC$_1$ algorithm. In this case, the difference in generalization skill may be attributed to the lagged correlation MLRs making predictions based on causally-spurious links. The causal relationship involved here can be understood in mass adjustment terms:mass conservation requires low-level convergence and upper-level divergence to be balanced by upward mass transport. This upward motion can invigorate convection and aid TC intensification. Hence, the vertical motion should be considered as a \textit{consequence} of divergence rather than an independent process that drives TC intensification \textit{by itself}. We believe that the removal of mid-level vertical motion in the PC$_1$ features shows that causal discovery algorithms can successfully remove causally-spurious links.
 \\
\section{Conclusion}
This paper described a causal feature selection framework to predict and understand complex geophysical events that can be considered multiple realizations of the same process with small perturbations. We applied this framework to multiple TC time series to identify common causal links and used them as predictors in MLR and RF regression models. Our results show that causal feature selection is superior to traditional feature selection methods for finding sets of predictors that help regression models generalize to unseen TC cases, especially for very sparse models (Fig~\ref{fig3}).  Of the two causal methods, we find that the PC$_1$ algorithm is more appropriate for feature selection, as it only keeps the most informative features, effectively removes confounding features (e.g., mid-tropospheric vertical motion in Fig~\ref{fig4}), and is less sensitive to the forecast lead time (Fig S3-S5) than PCMCI.\\
Temporally aligning the multiple time series based on a common reference point before causal feature selection tangibly improves model prediction skills. The retention of spurious links in the lag correlation models negatively affected generalizability. From these observations, we conclude that causal feature selection holds potential in our continued effort to improve statistical TC intensity models. Future efforts will involve (1) assessing whether current operational intensity prediction baselines can be improved by the causality-based predictor selection; (2) expanding the study to all ocean basins; and (3) discovering new potential predictors that may improve operational TC intensity predictions. \\
While not studied in this paper, the multidata causal discovery also opens the possibility to analyze systems whose causal structure changes in time: If one can align the individual member time series on a common time axis and can assume that, although changing in time, their causal structures are the same, then a dataset for independence testing can still be created by taking one sample per member time series. Repeating this procedure for every time step would yield one set of predictors per variable and per time step. Similarly, one could obtain one set of predictors per variable and per time window in a sequence of time windows (useful for slowly varying causal structures). Finally, we note that the multidata approach does not rely on any particular causal discovery algorithm. Therefore, while not shown here, the multidata approach can also be employed with the PCMCI$^+$ algorithm \cite{runge2020discovering}, a variant of PCMCI that allows contemporaneous causal influences and the Latent-PCMCI (LPCMCI) algorithm \cite{gerhardus2020high}, a variant of PCMCI that allows for contemporaneous causal influences and latent confounders (available within the open-source Python package \href{https://github.com/jakobrunge/tigramite}{tigramite}). Lastly, one could further optimize predictions by selecting the subset of causal predictors with the highest (validation set-)skill as discussed from an information-theoretic perspective in \cite{runge2015optimal}. In our context, however, iterating through all feature subsets is computationally prohibitive.

\paragraph{Acknowledgments}
We thank the DCSR at UNIL for providing the computational resources and technical support. 

\paragraph{Author Contributions}
Conceptualization: S.G.S.; T.B.; F.I.T., M.G  Methodology: J.R.; A.G; T.B., S.G.S. Data curation: S.G.S.; F.I.T.; T.B; Data visualization: S.G.S., F.I.T.; T.B. Writing original draft: S.G.S., All authors contributed to writing and review. All authors approved the revised manuscript.

\paragraph{Competing interest}
The authors declare that no competing interests exist. 

\paragraph{Data Availability Statement}
The codes and tutorials for multidata causal discovery are freely available in the \href{https://github.com/jakobrunge/tigramite}{Tigramite} GitHub repository and have been archived in Zenodo at https://doi.org/10.5281/zenodo.7747255. Sample code for the application are available at \href{https://github.com/saranya8989/Causal_ML}{Causal-ML} and have been archived in Zenodo at https://doi.org/10.5281/zenodo.7907217.. The WPAC TC data are from the \href{https://www.ncei.noaa.gov/products/international-best-track-archive}{IBtrACS} data archive. ERA5 datasets were downloaded from the Copernicus website (\href{https://cds.climate.copernicus.eu/cdsapp#!/dataset/reanalysis-era5-pressure-levels?tab=form}{multiple pressure levels} as well as \href{https://cds.climate.copernicus.eu/cdsapp#!/dataset/reanalysis-era5-single-levels?tab=form}{single pressure levels}).

 \paragraph{Ethics Statement}
 The research meets all ethical guidelines.
 
\paragraph{Funding Statement}
This research was supported by the canton of Vaud in Switzerland. J.R. has received funding from the European Research Council (ERC) Starting Grant CausalEarth under the European Union’s Horizon 2020 research and innovation program (Grant Agreement No. 948112).

\paragraph{Provenance}
This article is part of the Climate Informatics 2023 proceedings and was accepted in Environmental Data Science on the basis of the Climate
Informatics peer review process.

\paragraph{Supplementary Information}
Supplementary materials, organized into four sections, are provided with this manuscript.

\bibliographystyle{unsrt}  
\bibliography{main.bbl}  
\clearpage

\pagestyle{fancy}
\thispagestyle{empty}
\rhead{ \textit{ }}

\fancyhead[LO]{Feature selection using Multidata Causal Discovery}

\title{Supplementary Material: Selecting Robust Features for Machine Learning Applications using Multidata Causal Discovery
}

\author{
  Saranya Ganesh S., Tom Beucler, Frederick Iat-Hin Tam, Milton S. Gomez \\
  Institute of Earth Surface Dynamics \\
  University of Lausanne \\
  Lausanne, Vaud, Switzerland\\
   \And
  Jakob Runge \\ 
  1. Institute of Data Science \\
  German Aerospace Center (DLR) \\
  Jena, Thuringia, Germany \\
  2. Faculty of Electrical Engineering and Computer Science\\
  Technische Universität Berlin \\
  Berlin, Germany \\
  \And
  Andreas Gerhardus \\
  Institute of Data Science \\
  German Aerospace Center (DLR) \\
  Jena, Thuringia, Germany\\
}

\maketitle

\setcounter{page}{1}

In Section A, we included the full set of variables for our prediction problem. Section B of this supplemental information shows the results of the performance of causal models for the remaining targets, including Minimum Sea Level Pressure (MSLP) and Total Integrated Precipitation. Section B also includes results for experiments with a reduced lead time of 6 hours, using both the PC1 and PCMCI methods. The feature selection baselines and details of ML models used for comparing the performances are defined in Section C, followed by the results for predicting the remaining targets in Section D. Finally, Section E shows the performance of the best models, as well as the causal predictors used in the model with the best skill on the validation set for maximum surface wind. \\

\section*{A. List of Variables Used as Predictors} \label{secA}
We provide a list of all the variables chosen from the ERA5 (3 hourly) dataset, including targets and predictors, for preparing the ensemble of TC time series in Table~\ref{tab:vars}.

\begin{table}[htb]
\centering
{\includegraphics[width=1.\textwidth]{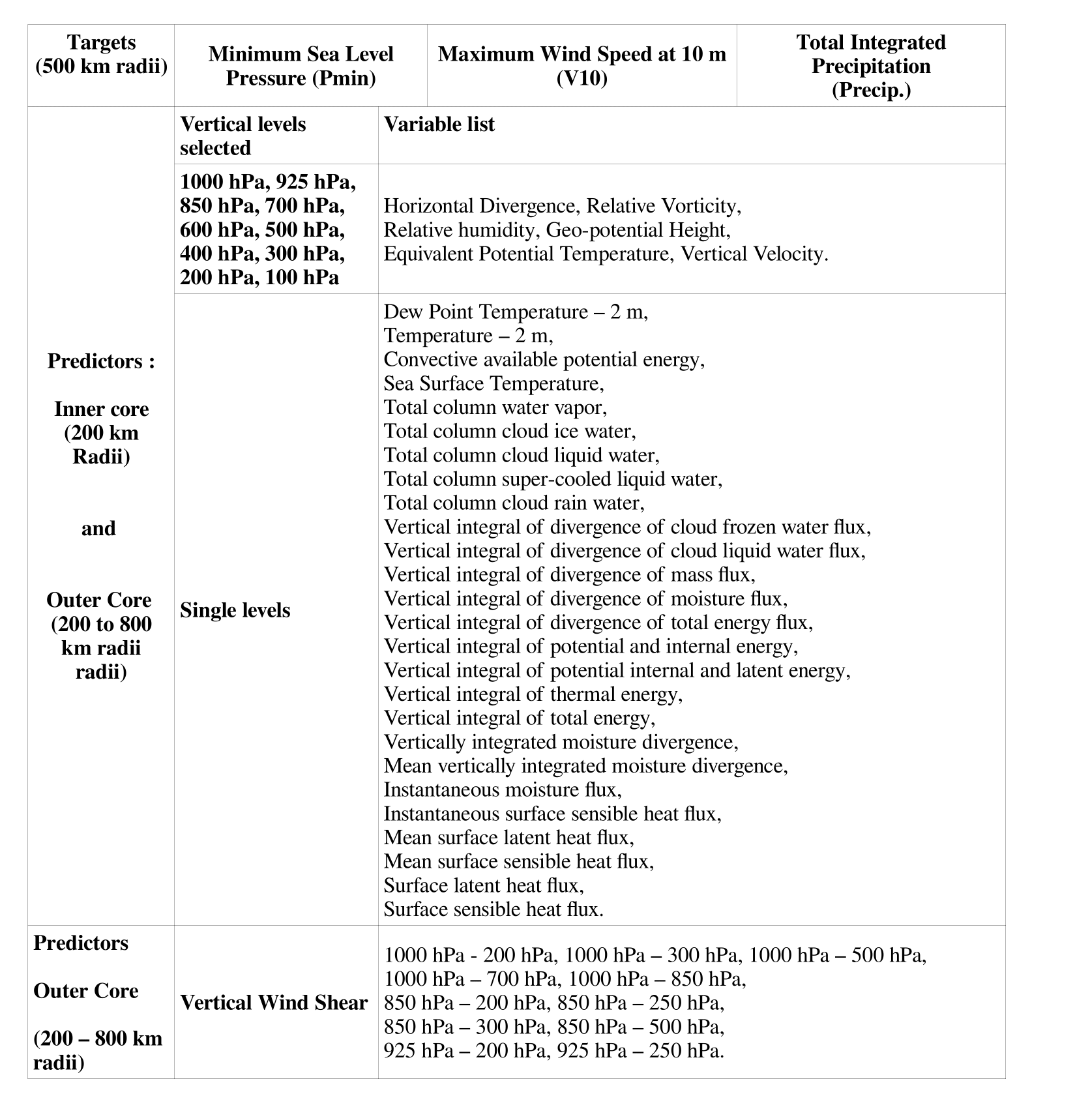}}
{\caption{List of variables used from ERA5 dataset.}
\label{tab:vars}}
\end{table}

\section*{B. Optimal Number of Causal Features} \label{secB}

\begin{figure}[htb]%
     \centering
    \includegraphics[width=1\textwidth]{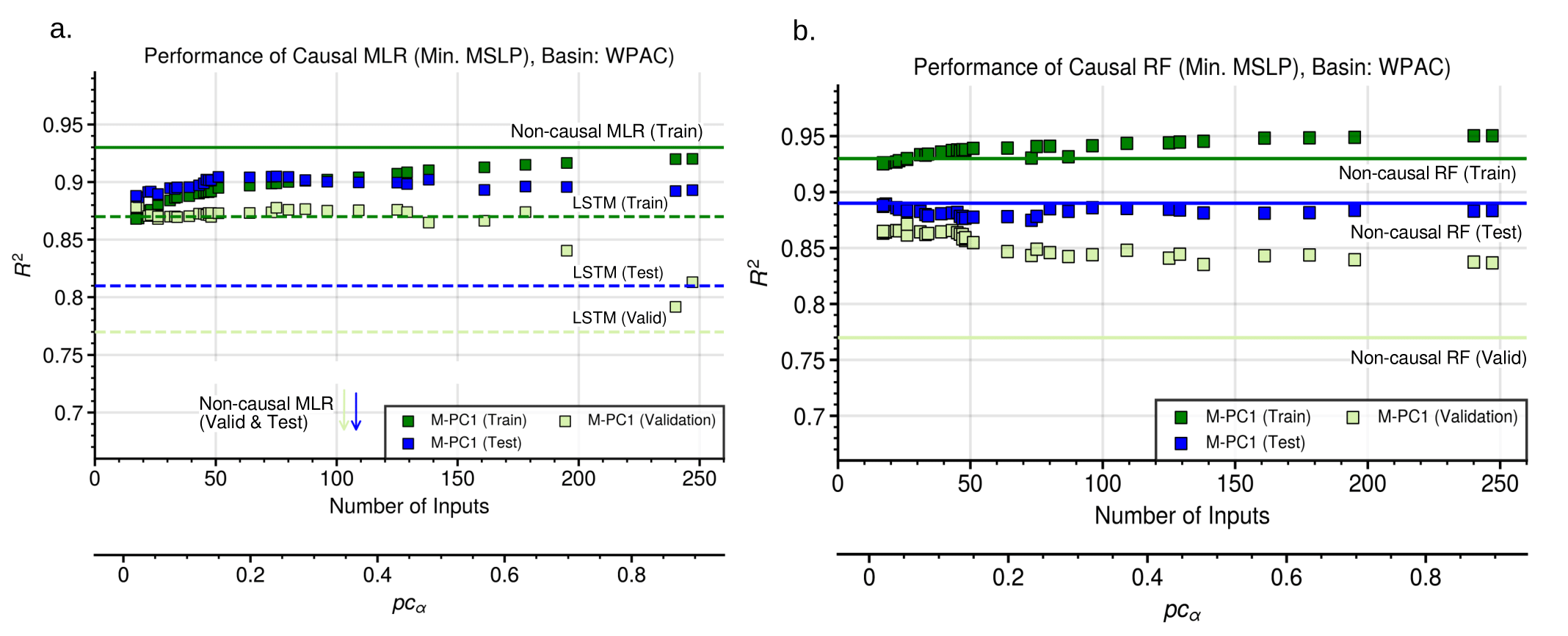}
    {\caption{Performance of Causal ML for multiple tests by varying hyperparameter for the prediction of Minimum MSLP by causal-MLR (a) compared to noncausal ML (Solid lines) and LSTM (dashed lines), where as (b) shows the performance of Causal-RF compared to noncausal-RF (solid lines)}
    \label{si:fig1}}
\end{figure}

\begin{figure}[htb]%
    \centering
    \includegraphics[width=1.\textwidth]{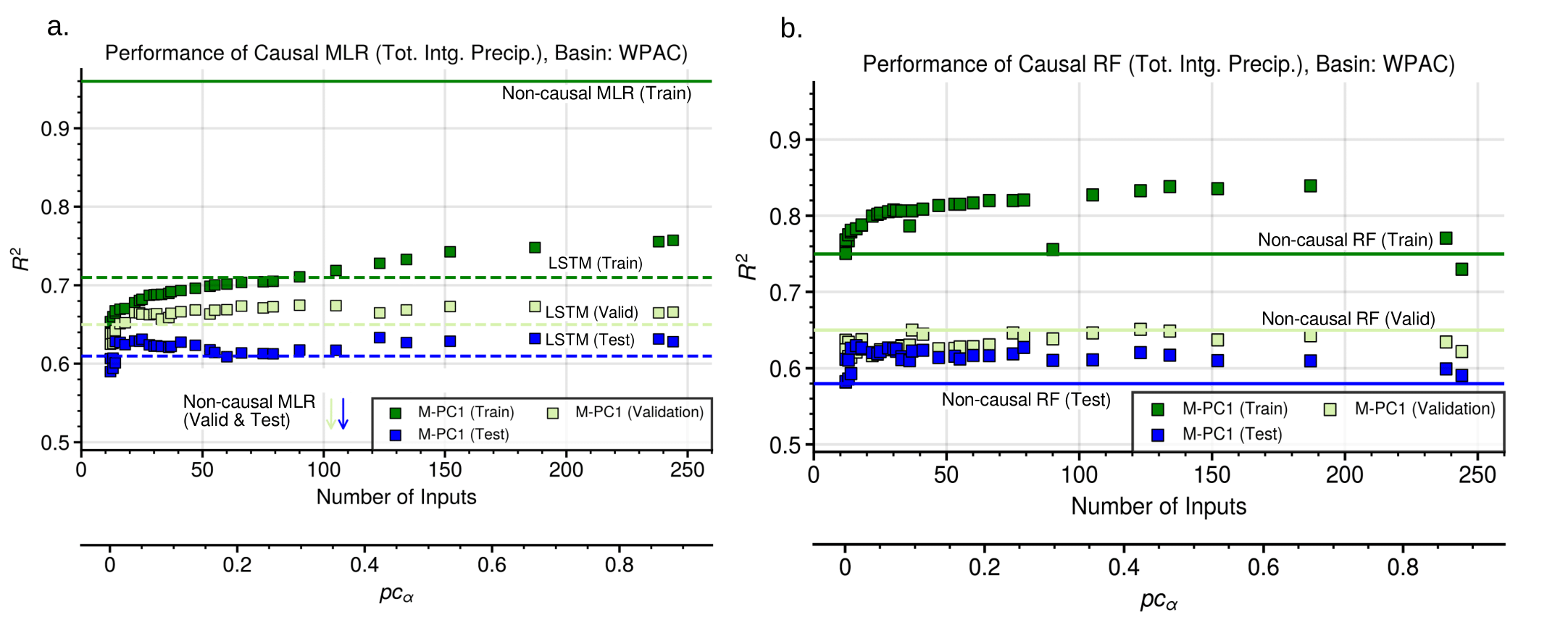}
    {\caption{Same as Figure 1 but for the prediction of Total integrated Precipitation}
    \label{si:fig2}}
\end{figure}

\begin{figure}[htb]%
    \centering
    \includegraphics[width=1\textwidth]{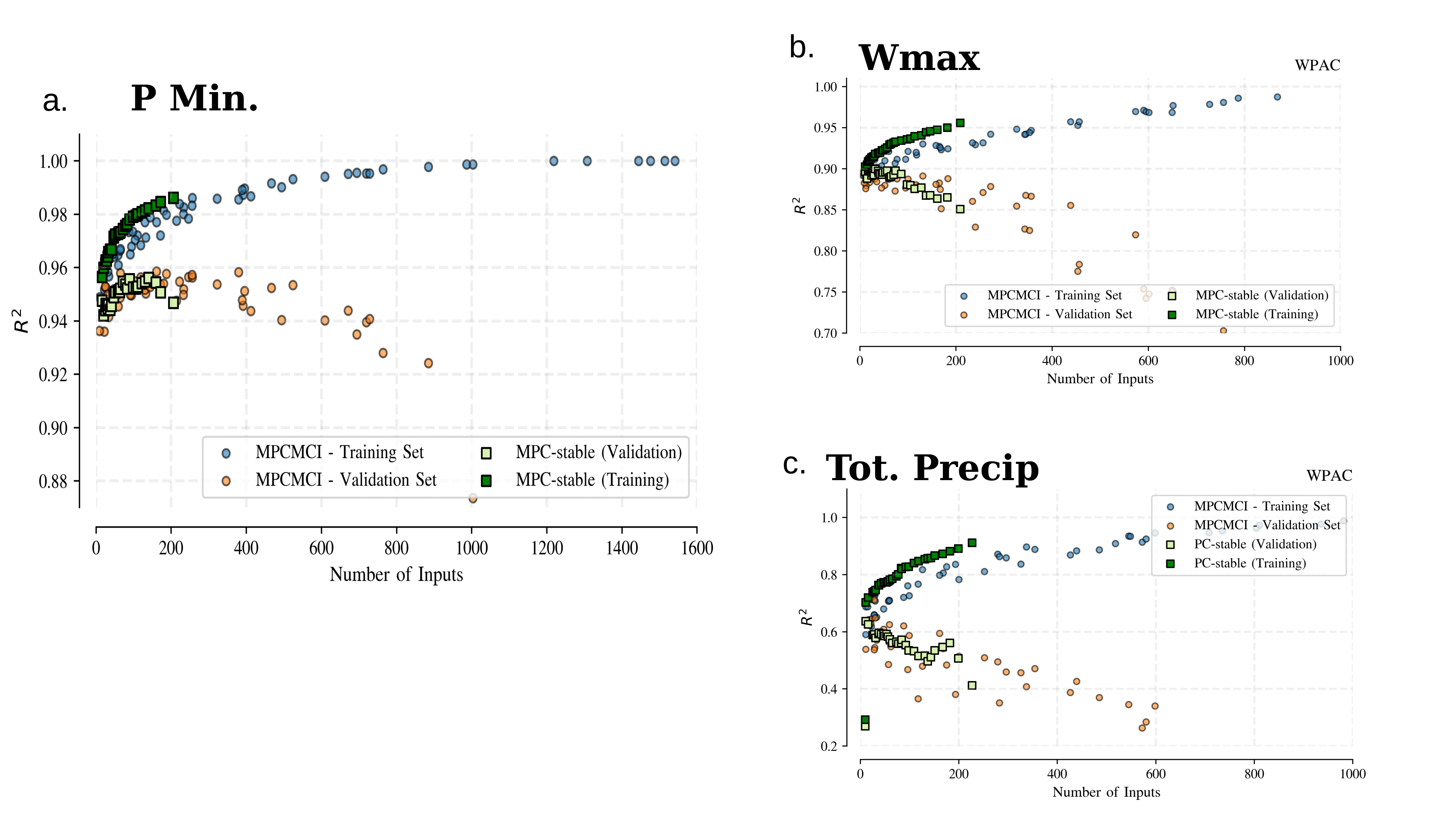}
    {\caption{Comparison of M-$PC_{1}$ and M-PCMCI based Causal-MLR model performance for 6 hour predictions without time alignment and with time lags ranging from 6 hrs to 2 days for selected targets}
    \label{si:fig3}}
\end{figure}

\begin{figure}[htb]%
    \centering
    \includegraphics[width=1\textwidth]{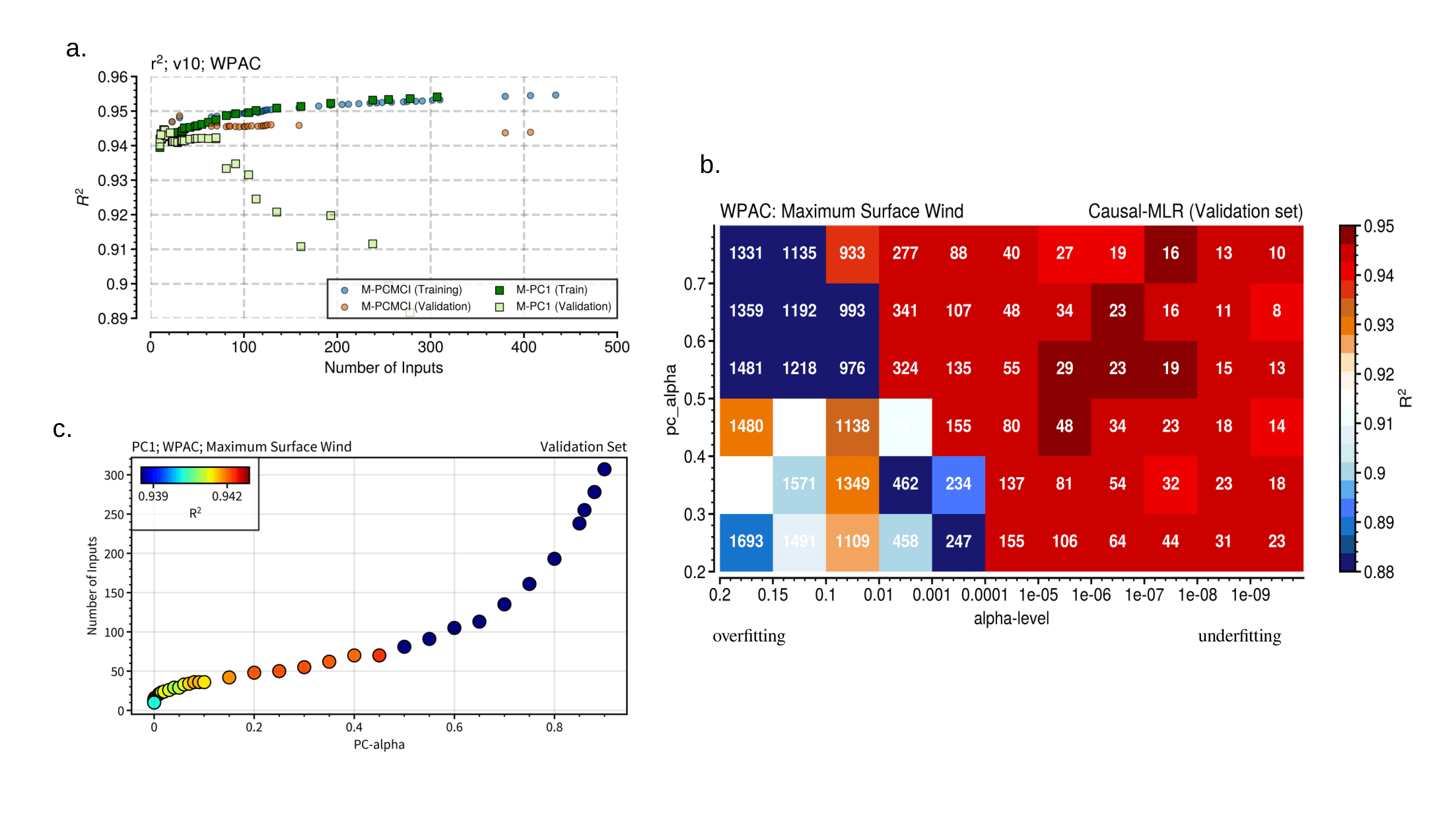}
    {\caption{Performance of $PC_{1}$ and PCMCI models with time aligned inputs for the prediction of Maximum Wind (a), the relationship between hyper-parameters, inputs, and performance (b, c)}
    \label{si:fig4}}
\end{figure}

\begin{figure}[htb]%
\centering
    \includegraphics[width=1.1\textwidth]{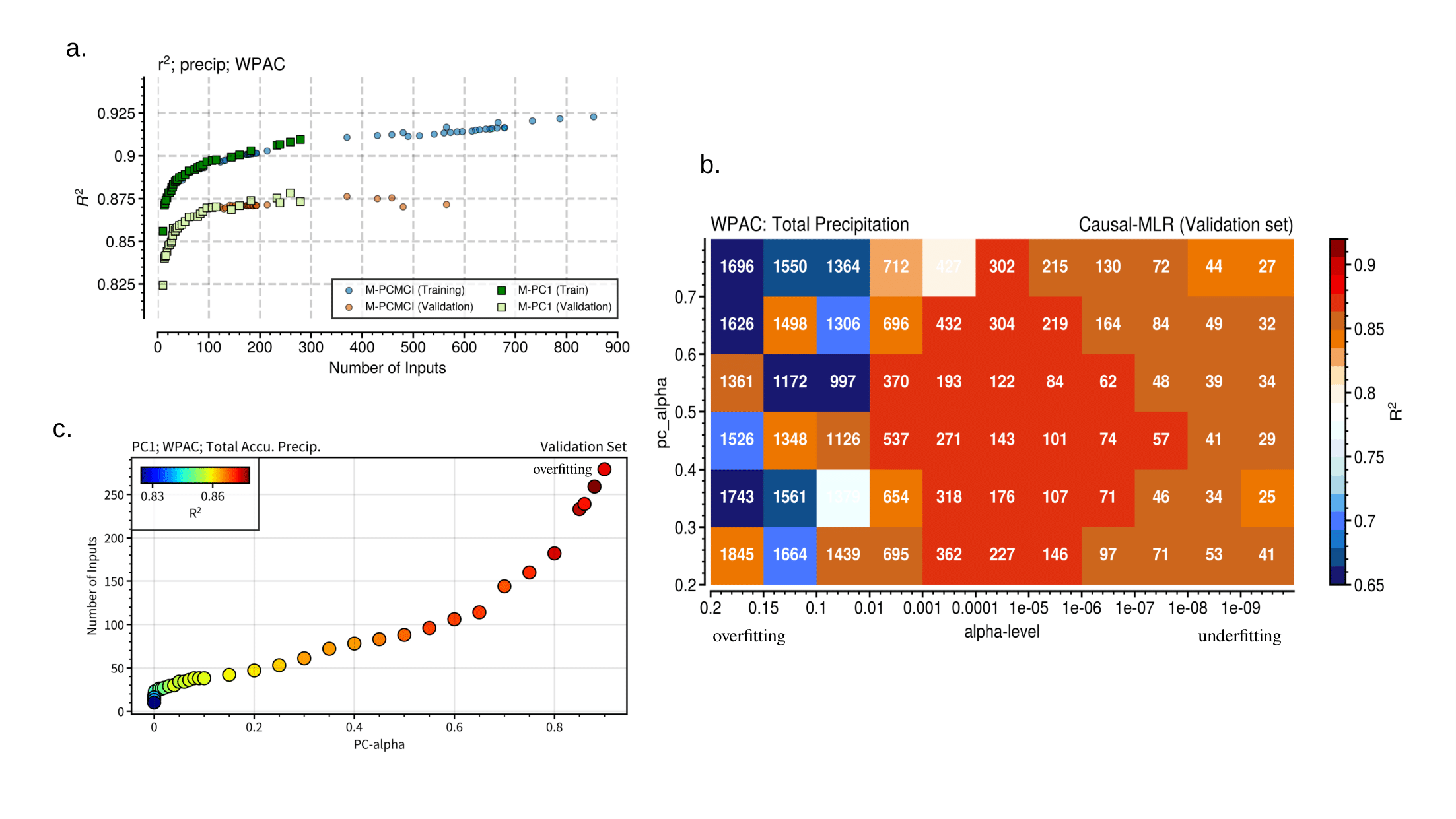}
    {\caption{Same as the previous figure, but for the prediction of total integrated precipitation}
    \label{si:fig5}}
\end{figure}

\begin{figure}[htb]%
    \centering
    \includegraphics[width=1.1\textwidth]{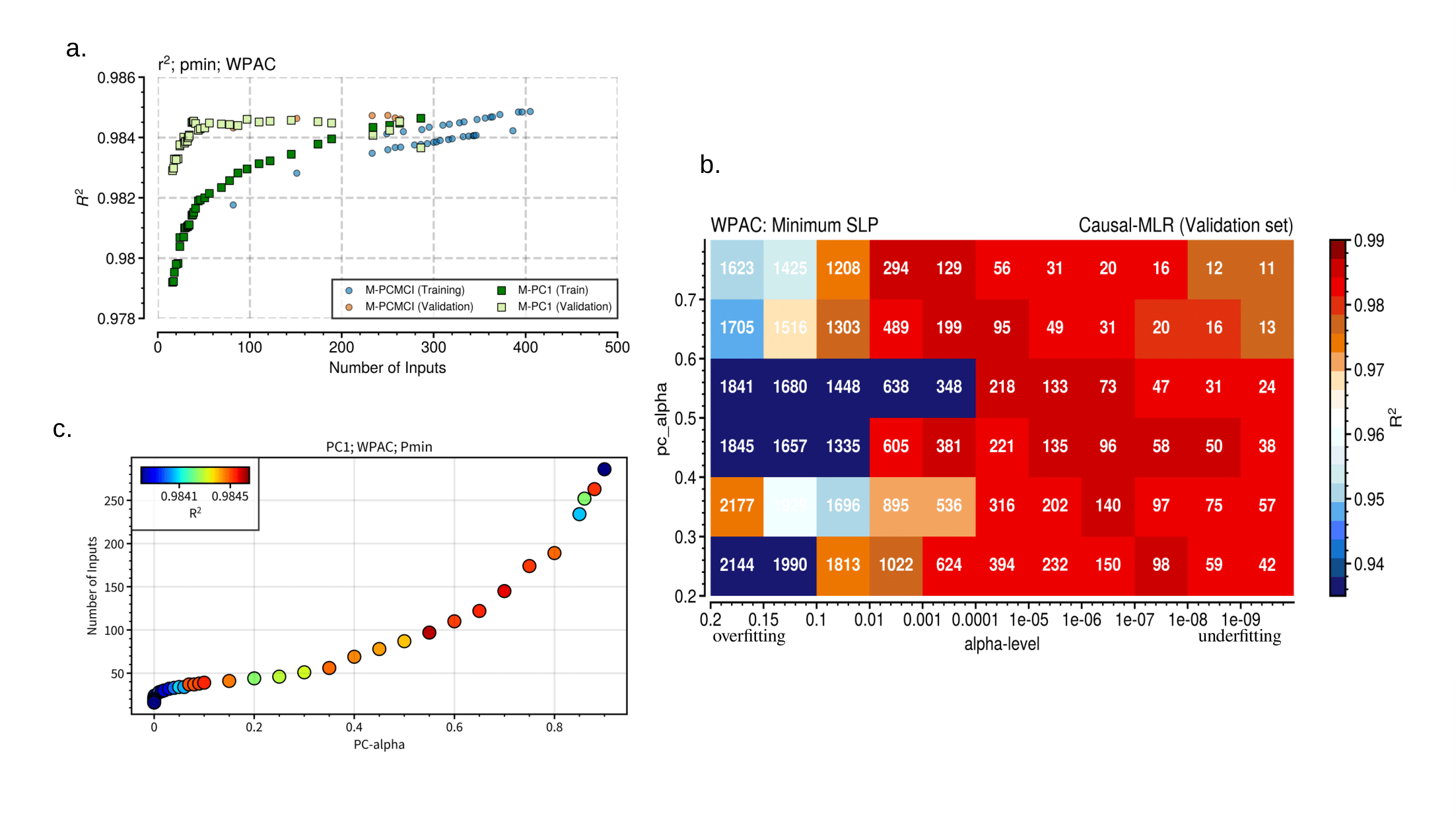}
    {\caption{Same as the previous figure, but for the prediction of Minimum MSLP}
    \label{si:fig6}}
\end{figure}

 Figures \ref{si:fig3},\ref{si:fig4},\ref{si:fig5}, \& \ref{si:fig6} show the comparison of M-PC$_{1}$ and M-PCMCI algorithms for the selected targets before and after temporally aligning the time series according to the time of minimum MSLP during the lifetime of each storms in the group. We see a clear improvement in the validation sets of the aligned dataset for both PC1 and PCMCI for all the targets.

\section*{C. Description of Machine Learning Algorithms} \label{secC} %

In this section, we provide a description of our implementations of the machine learning algorithms tested for this work. Prior to training the algorithms, we calculated the mean and standard deviation of each input feature in the available training data, noting that the values were considered for the set of all storms (and not on a per-storm basis). We then used these values to \textit{standard-scale} the input features, per Eqn. (\ref{eq:scaling}).  
\begin{equation}
    z_{i} = \dfrac{x_i - \mu_i}{\sigma_i}
    \label{eq:scaling}
\end{equation}

\textbf{Multiple linear Regression} - In order to benchmark the performance of multidata causal feature selection, a plurality of multiple linear regression (MLR) algorithms were prepared, using the Scikit-Learn implementation of the Linear Regression algorithm and its corresponding default parameters. Each individual MLR algorithm was trained to predict one of three unscaled target variables (i.e., one of MSLP, precipitation, or Surface Wind) using the selected, standard-scaled inputs being evaluated.

\textbf{Random Forest Regression} - To ensure that the benefit of causal feature selection extends to more complex, nonlinear machine learning algorithms. We applied the same sets of input variables used to train the causal and non-causal MLR models to a Random Forest Regressor (RF Regressor). The implementation of the RF regression algorithm in this study utilizes that provided in the Scikit-Learn package. Compared to the MLR models, the RF Regressor contains several trainable hyperparameters that we can optimize for better prediction skills. Using the \textit{RandomizedSearchCV} function, we tuned the hyperparameters related to the depth of the model, minimum number of samples to split decision trees, and the number of estimators. The best model that has the best cross-validation accuracy on the training data is chosen for analysis.

The noncausal feature selection baselines that are used in the main manuscript are described below.

\textbf{Random Feature Selection} is a sampling method where features are chosen randomly. Random sampling is analogous to drawing out a set of cards after shuffling without any criteria. Our implementation of this algorithm randomly selects a set of input features (size ranging from 10 to 1000) from all possible combinations of variables and time lags. 

\textbf{Lagged Correlation} considers the absolute correlation between the prediction targets and different time-lagged input features. We adopted a kitchen sink approach where we obtained the correlation values between targets and all time-lagged variables by c. These correlation values are ranked and the features with the highest correlations are then chosen as MLR inputs. The size of these sets of features ranges from 10 to 1000.

\textbf{XAI} takes the training dataset to build a random forest regression model using Python's scikit-learn library. By using this baseline method, we explore whether the use of feature importance (when nonlinear relationships between variables and targets are included) can result in a better selection of features. The Gini feature importance as measured by the trained random forest regressors provides an objective means to rank and select the most informative input variables. Input variables are ranked from most important to least important based on Gini impurity-based feature importances. The top-ranked features are then chosen to train the MLR models. Alternative feature importance methods, e.g., permutation feature importance or absolute Shapley values, are left for future work.

\textbf{LSTM Neural Network} - We prepared three Long Short-Term Memory (LSTM) recurrent neural networks as baselines, training the LSTM models on standard-scaled input data and configuring each LSTM to predict one of the standard-scaled target variables (i.e., one of MSLP, precipitation, or Surface Wind). We implemented each LSTM as a sequential model using PyTorch; their architecture includes an LSTM layer, a dropout layer, a linear hidden layer, and a linear output layer. As we targeted standard scaled outputs, the output of the network needed to represent positive and negative values. To do this, we set the output activation function to the identity function and we set the hidden layer activation function to hyperbolic tangent. We selected the Adam optimizer and mean-square error loss for our training, and proceeded to conduct a hyperparameter search using the Optuna framework. The study employed 10 trials that tested LSTM and hidden layers 50-100 units wide, dropout rates between 0.0 and 0.5, and learning rates between $1\mathrm{e}{-4}$ and $1\mathrm{e}{-3}$. We note that we set the number of units in the LSTM layer and the hidden layer to be equal to each other in all conducted trials.

\section*{D. Comparison of Feature Selection Baselines} \label{secD}

\begin{figure}[htb]
    \centering
    \includegraphics[width=0.9\textwidth]{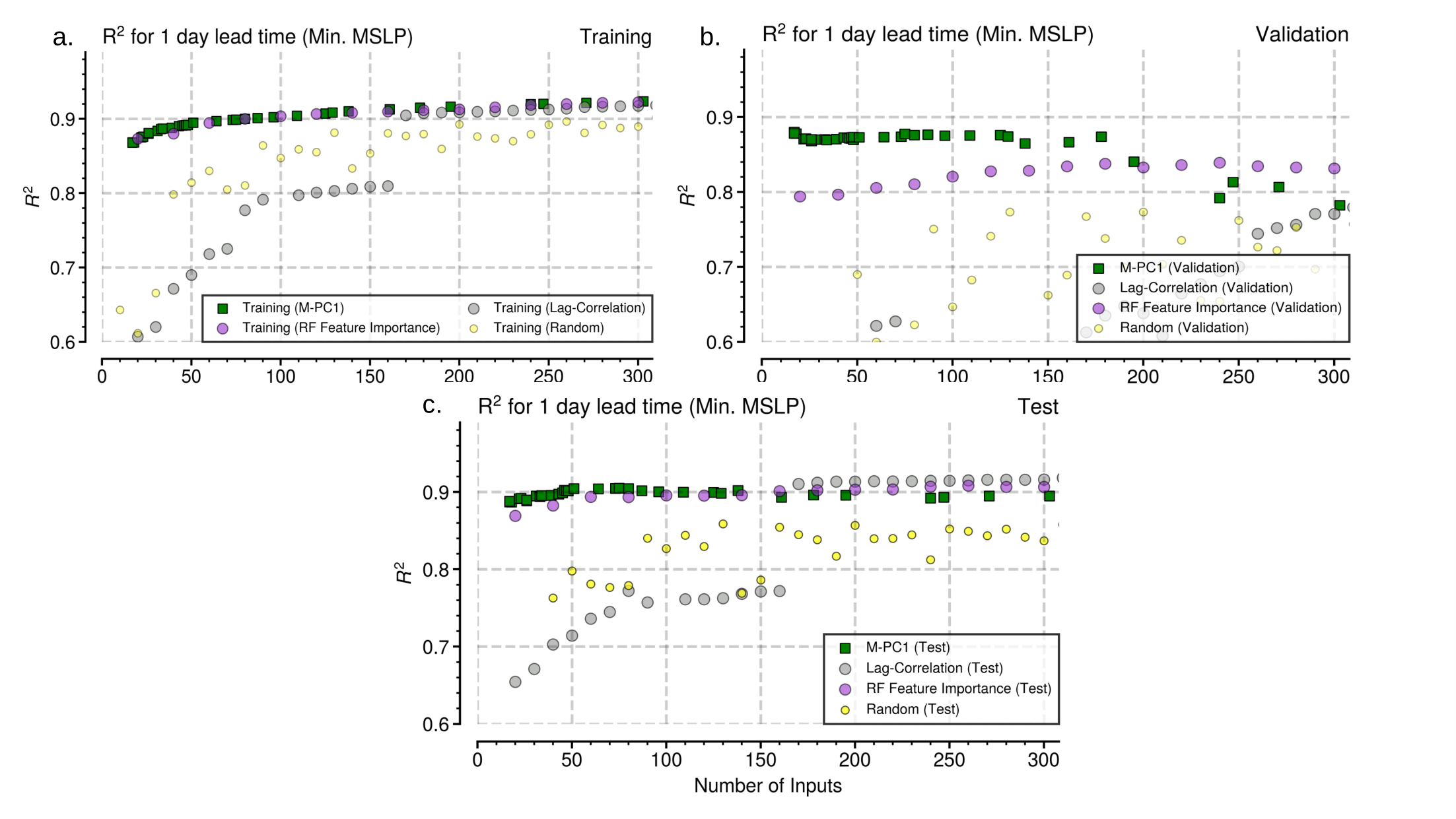}
    \caption{Comparison of the performance of Training, Validation and Test sets of MLR models that used different feature selection methods for predicting minimum MSLP}
    \label{si:fig7}
 \end{figure}

\begin{figure}[htb]%
    \centering
    \includegraphics[width=0.9\textwidth]{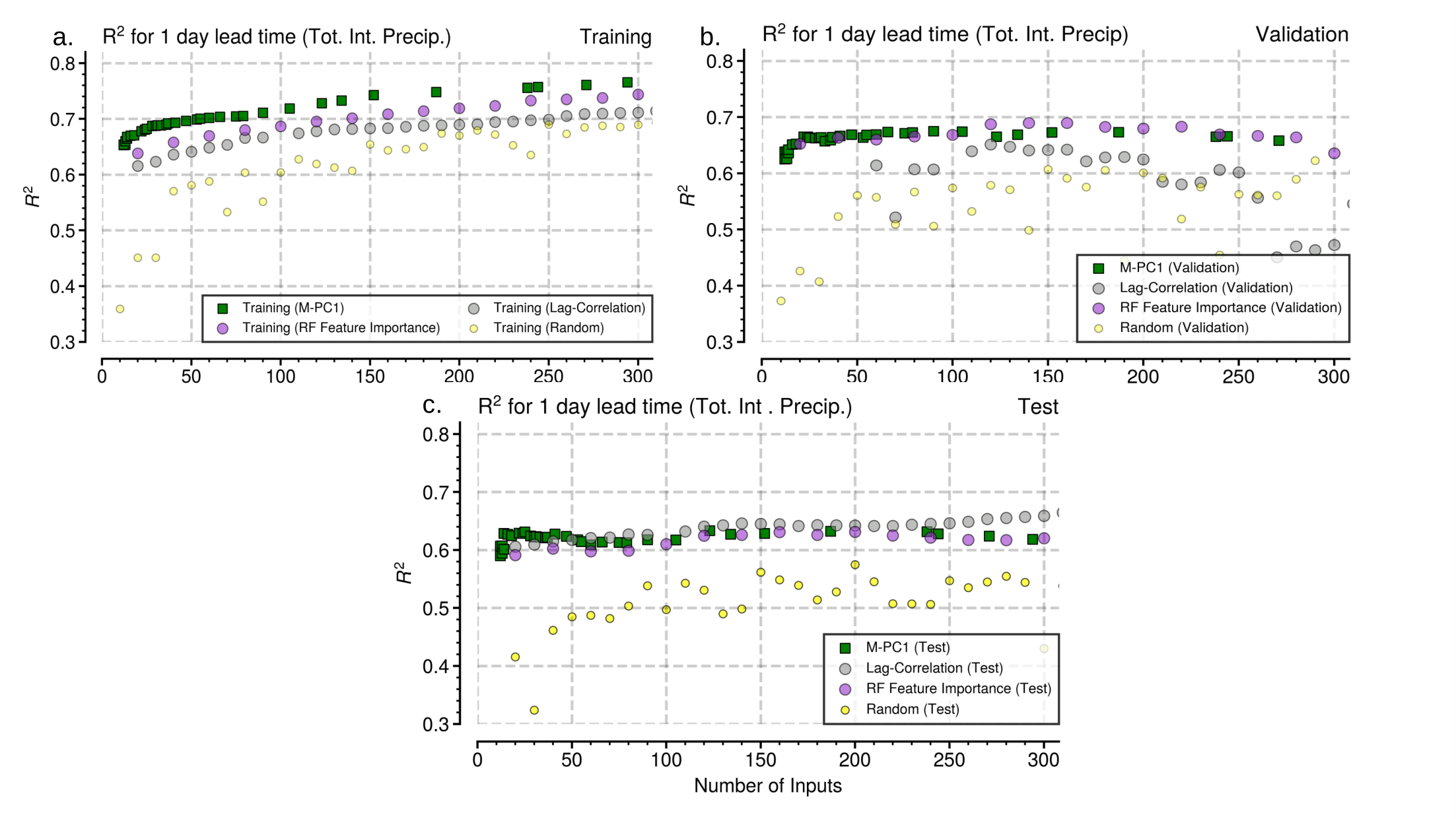}
    \caption{Same as Figure 7. but for predicting Total area integrated precipitation}
    \label{si:fig8}
 \end{figure}

A comparison of the performance of Causal MLR to the performance of MLR models based on other feature selection baselines for the targets, Minimum MSLP and Total Integrated Precipitation are shown in Figures \ref{si:fig7} and \ref{si:fig8} respectively.

The Root Mean Square Error (RMSE) and Mean Absolute Error (MAE) of the best model prediction of \emph{Maximum Surface Winds, MSLP and Total integrated Precipitation} on both the training, validation and test sets for the best ML models. All metrics signify a good performance for Causal-ML with far less number of inputs compared to the number of inputs from the best models using the Non-causal-RF and Non-causal MLR methods. For the best ML models used, RMSE are listed in Table \ref{tab:rmse} and MAE are listed Table \ref{tab:mae}.

\section*{E. Optimal Causal Predictors} \label{secE}
The predictors and time lags for the best causal-MLR model with time-aligned inputs for the prediction of maximum wind speeds 1-day in advance are shown in Table \ref{tab:causal}.

\begin{table}
\begin{centering}
\resizebox{\textwidth}{!}{%
\begin{tabular}{|c|c|c|c|c|c|c|c|c|c|c|}
\hline 
\multicolumn{2}{|c|}{ML Models} & \multicolumn{3}{c|}{Training\textcolor{purple}{{} }\textit{\textcolor{purple}{(No. of features)}}} & \multicolumn{3}{c|}{Validation} & \multicolumn{3}{c|}{Test }\tabularnewline
\hline 
\multicolumn{2}{|c|}{\textbf{Target}} & Pmin ($hPa$)\textcolor{purple}{{} } & V10 ($ms^{-1}$) & Precip $\times 10^{-3}$ ($km^{2}$) & Pmin \textcolor{purple}{{} } & V10 & Precip & Pmin \textcolor{purple}{{} } & V10 & Precip \tabularnewline
\hline 
\multicolumn{2}{|c|}{\textbf{Causal RF}} & 13.45 \textit{\textcolor{purple}{ (26)}} & 3.49 \textit{\textcolor{purple}{ (17)}} & 35.45 \textit{\textcolor{purple}{ (123)}} & 28.05 & 6.24 & 63.3 & 24.03 & 6.26 & 92.1\tabularnewline
\hline 
\multicolumn{2}{|c|}{\textbf{Causal MLR}} & 24.67 \textit{\textcolor{purple}{ (17)}} & 5.2 \textit{\textcolor{purple}{ (31)}} & 61.38 \textit{\textcolor{purple}{ (90)}} & 25.3 & 5.62 & 59.01 & 21.89 & 5.87 & 92.79\tabularnewline
\hline 
\multirow{3}{*}{\textbf{Non-causal RF}} & All & \textcolor{black}{15.68}\textit{\textcolor{purple}{{} (3978)}} & \textcolor{black}{3.91}\textit{\textcolor{purple}{{} (3978)}} & \textcolor{black}{46.25}\textit{\textcolor{purple}{{} (3978)}} & 34.92 & 7.58 & 100.03 & 24.39 & 6.54 & 102.16\tabularnewline
 & Lagged & \textcolor{black}{8.80}\textit{\textcolor{purple}{{} (480)}} & \textcolor{black}{2.3}\textit{\textcolor{purple}{{} (560)}} & \textcolor{black}{37.8}\textit{\textcolor{purple}{{} (80)}} & 29.22 & 6.23 & 82.12 & 21.04 & 5.65 & 93.2\tabularnewline
 & Random & \textcolor{black}{8.57}\textit{\textcolor{purple}{{} (870)}} & \textcolor{black}{2.27}\textit{\textcolor{purple}{{} (770)}} & \textcolor{black}{27.94}\textit{\textcolor{purple}{{} (970)}} & 38.51 & 7.8 & 105.23 & 27.15 & 7.19 & 105.77\tabularnewline
\hline 
\multirow{4}{*}{\textbf{Non-causal MLR}} & All & \textcolor{black}{2.43 }\textit{\textcolor{purple}{(3978)}} & \textcolor{black}{0.66}\textit{\textcolor{purple}{{} (3978)}} & \textcolor{black}{7.87}\textit{\textcolor{purple}{{} (3978)}} & 373.54 & 398.33 & 33370.94 & 105.74 & 31.03 & 338.29\tabularnewline
 & Lagged & \textcolor{black}{15.64}\textit{\textcolor{purple}{{} (440)}} & \textcolor{black}{12.04}\textit{\textcolor{purple}{{} (40)}} & \textcolor{black}{60.60}\textit{\textcolor{purple}{{} (120)}} & 31.09 & 14.77 & 91.40 & 17.20 & 11.83 & 85.70\tabularnewline
 & Random & \textcolor{black}{19.05}\textit{\textcolor{purple}{{} (420)}} & \textcolor{black}{5.58}\textit{\textcolor{purple}{{} (130)}} & \textcolor{black}{58.23}\textit{\textcolor{purple}{{} (290)}} & 41.87 & 8.20 & 97.68 & 29.88 & 7.81 & 110.5\tabularnewline
 & XAI & 16.93\textit{\textcolor{purple}{{} (240)}} & 3.84\textit{\textcolor{purple}{(420)}} & 55.34\textit{\textcolor{purple}{{} (140)}} & 30.90 & 6.05 & 80.32 & 19.92 & 6.38 & 90.7\tabularnewline
\hline 
\multicolumn{2}{|c|}{\textbf{LSTM}} & 27.55 \textit{\textcolor{purple}{ (3978)}} & 6.49 \textit{\textcolor{purple}{(3978)}} & 179.76 \textit{\textcolor{purple}{ (3978)}} & 44.00 & 8.80 & 199.29 & 39.44 & 8.02 & 206.12\tabularnewline
\hline 
\end{tabular}}
\caption{RMSE.\label{tab:rmse}}
\par\end{centering}
\end{table}

\begin{table}
\begin{centering}
\resizebox{\textwidth}{!}{%
\begin{tabular}{|c|c|c|c|c|c|c|c|c|c|c|}
\hline 
\multicolumn{2}{|c|}{ML Models} & \multicolumn{3}{c|}{Training\textcolor{purple}{{} }\textit{\textcolor{purple}{(No. of features)}}} & \multicolumn{3}{c|}{Validation} & \multicolumn{3}{c|}{Test }\tabularnewline
\hline 
\multicolumn{2}{|c|}{\textbf{Target}} & Pmin ($hPa$)\textcolor{purple}{{} } & V10 ($ms^{-1}$) & Precip $\times 10^{-3}$ ($km^{2}$) & Pmin \textcolor{purple}{{} } & V10 & Precip & Pmin \textcolor{purple}{{} } & V10 & Precip \tabularnewline
\hline 
\multicolumn{2}{|c|}{\textbf{Causal RF}} & 2.55 \textit{\textcolor{purple}{(26)}} & 1.42 \textit{\textcolor{purple}{(17)}} & 0.14 \textit{\textcolor{purple}{(123)}} & 3.87 & 1.94 & 0.19 & 3.62 & 1.97 & 0.23\tabularnewline
\hline 
\multicolumn{2}{|c|}{\textbf{Causal MLR}} & 3.49\textit{\textcolor{purple}{(17)}} & 1.77 \textit{\textcolor{purple}{(31)}} & 0.19 \textit{\textcolor{purple}{(90)}} & 3.62 & 1.84 & 0.19 & 3.49 & 1.89 & 0.23\tabularnewline
\hline 
\multirow{3}{*}{\textbf{Non-causal RF}} & All & 2.81 \textit{\textcolor{purple}{(3978)}} & 1.50 \textit{\textcolor{purple}{(3978)}} & 0.16 \textit{\textcolor{purple}{(3978)}} & 4.11 & 2.17 & 0.23 & 3.65 & 1.99 & 0.24\tabularnewline
 & Lagged & 2.04 \textit{\textcolor{purple}{(480)}} & 1.12 \textit{\textcolor{purple}{(560)}} & 0.14 \textit{\textcolor{purple}{(80)}} & 3.78 & 1.95 & 0.22 & 3.41 & 1.85 & 0.23\tabularnewline
 & Random & 2.07 \textit{\textcolor{purple}{(870)}} & 1.12 \textit{\textcolor{purple}{(770)}} & 0.12 \textit{\textcolor{purple}{(970)}} & 4.43 & 2.2 & 0.25 & 3.94 & 2.1 & 0.25\tabularnewline
\hline 
\multirow{4}{*}{\textbf{Non-causal MLR}} & All & 1.21 \textit{\textcolor{purple}{(3978)}} & 0.63 \textit{\textcolor{purple}{(3978)}} & 0.07 \textit{\textcolor{purple}{(3978)}} & 10.42 & 6.15 & 0.85 & 7.71 & 4.44 & 0.44\tabularnewline
 & Lagged & 2.90 \textit{\textcolor{purple}{(440)}} & 1.54 \textit{\textcolor{purple}{(40)}} & 0.17 \textit{\textcolor{purple}{(120)}} & 3.74 & 2.07 & 0.22 & 3.11 & 1.70 & 0.21\tabularnewline
 & Random & 3.37\textit{\textcolor{purple}{{} (420)}} & 1.56 \textit{\textcolor{purple}{(130)}} & 0.19 \textit{\textcolor{purple}{(290)}} & 5.11 & 2.32 & 0.24 & 4.14 & 2.17 & 0.24\tabularnewline
 & XAI & 3.03 \textit{\textcolor{purple}{(240)}} & 1.76 \textit{\textcolor{purple}{(420)}} & 0.17 \textit{\textcolor{purple}{(140)}} & 4.07 & 1.97 & 0.22 & 3.4 & 1.93 & 0.22\tabularnewline
\hline 
\multicolumn{2}{|c|}{\textbf{LSTM}} & 3.90 \textit{\textcolor{purple}{(3978)}} & 1.97 \textit{\textcolor{purple}{(3978)}} & 0.34 \textit{\textcolor{purple}{(3978)}} & 4.85 & 2.29 & 0.35 & 4.74 & 2.22 & 0.36\tabularnewline
\hline 
\end{tabular}}
\caption{MAE (lower values means better models)\label{tab:mae}}
\par\end{centering}
\end{table}

\begin{table}[htb]
\centering
{\includegraphics[width=1.0\textwidth]{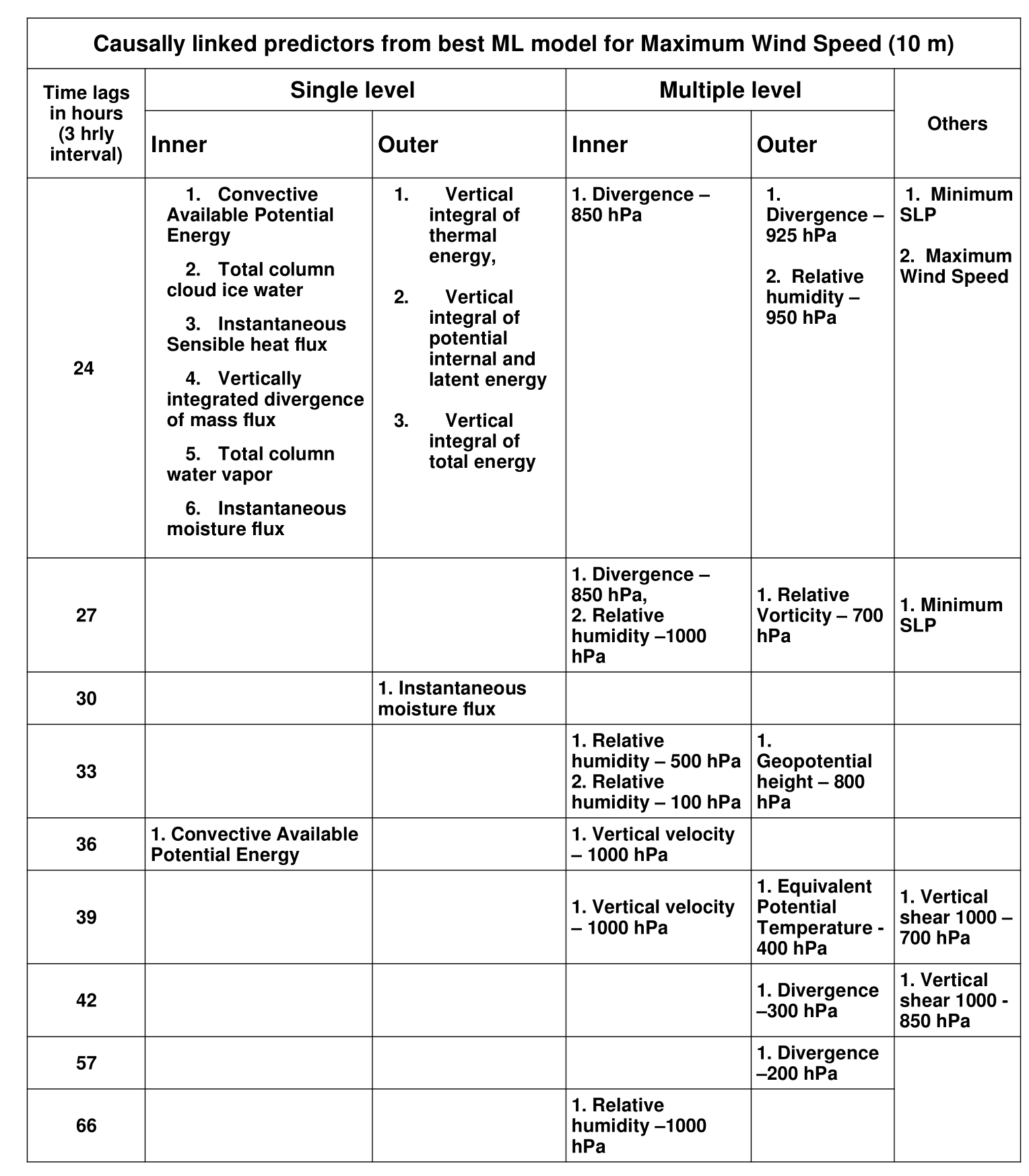}}
{\caption{List of 31 causally linked predictors for Maximum wind (1 day lead-time) at significant time lags with best model using $PC_{1}$}
\label{tab:causal}}
\end{table}



\end{document}